\documentclass[journal]{IEEEtran}
\usepackage{fontawesome5}
\usepackage[T1]{fontenc}
\usepackage[utf8]{inputenc}
\usepackage{amsmath,amssymb,amsfonts}
\usepackage{graphicx}
\usepackage{booktabs}
\usepackage{multirow}
\usepackage{array}
\usepackage{algorithm}
\usepackage{algorithmic}
\usepackage{textcomp}
\usepackage{xcolor}
\usepackage{url}
\usepackage{microtype}
\usepackage[hidelinks]{hyperref}
\usepackage{hyperref}
\graphicspath{{figures/}}
\newcommand{\SCARresult}[2]{\shortstack[c]{#1\\{\footnotesize(#2)}}}

\begin{document}

\title{SCAR: Bridging Coarse Retrieval and Fine Localization through Candidate Reranking}

\author{
Shuhan Dong
and Zhenyu Li*\\[6pt]
Shandong Academy of Sciences (SAC)\\
[6pt]
\href{https://github.com/shuhandong748-oss/scar}{
    \faGithub\ \textcolor{blue}{\texttt{github.com/shuhandong748-oss/scar}}
}
}


\markboth{CV4RA-LAB}%
{SCAR: Bridging Coarse Retrieval and Fine Localization through Candidate Reranking}

\maketitle
\begingroup
\renewcommand{\thefootnote}{}
\footnotetext{
$*$ Corresponding author.
}
\endgroup
\begin{abstract}
Natural-language point-cloud localization typically retrieves candidate map cells and then predicts a position inside each candidate, yet the final output is still determined by retrieval order. A retrieval score measures region-description relevance, not whether the coordinates a region yields are accurate: a relevant region may produce a displaced prediction, and a lower-ranked region may contain a successful one. We study the final selection problem after the candidate pool and Fine predictions are fixed---how to choose the coordinate most likely to localize successfully, and under which conditions the gain holds. We propose SCAR (Spatial Consistency-Aware Reranking). SCAR treats each cell--Fine-coordinate pair as a localization hypothesis, interprets directional relations as functions of position, and reads spatial support at the backbone's actual Fine point (Fine-conditioned Evidence Readout). It organizes this readout with center support, semantic presence, retrieval score, and geometric offsets into a 23-D representation, trains a shared scorer with localization-success labels, and at inference reranks existing hypotheses and returns their original coordinates. Experiments cover four public backbones and a corrected Des4Pos reimplementation across five domains. Excluding one protocol-reference configuration, exploratory TEST point estimates on the remaining 24 configurations show Top-1@5\,m above Original on 23 and Cell R@1 above Original on all 24. On development-set VAL, the Fine readout improves Top-1@5\,m by 5.77 percentage points (pp) over a center surrogate on corrected Des4Pos$\times$KITTI, and the complete scoring scheme improves it by 13.98\,pp over FSR on MambaPlace$\times$nuScenesFull; the latter measures the complete scheme, not learned ranking alone.
\end{abstract}


\section{Introduction}
Language-driven map localization converts a person's description of an environment into an executable location, providing an interface for specifying places without metric coordinates in shuttle, delivery, and human--robot collaboration scenarios. Unlike 3-D language grounding that points to a target object, a place description usually constrains an observation position through several objects and their relative directions: ``east of the gray lamp post and west of the black vegetation'' constrains a position among these objects, not any object itself. ScanRefer studies 3-D localization of the object referred to by language~\cite{chen2020scanrefer}, and Text2Pos established the task and a coarse-to-fine framework for text-based localization on urban point-cloud maps~\cite{kolmet2022text2pos}. In this task, language semantics, object layout, and the final output coordinate must all hold simultaneously.

The coarse-to-fine framework first retrieves candidate regions and then predicts a position in the local map, decomposing global search and local estimation into two stages. RET, Text2Loc/Text2Loc++, MNCL, MambaPlace, and CMMLoc improve representations and candidate quality through relation modeling, cross-modal alignment and fine localization, negative constraints, sequence interaction, and directional priors, respectively~\cite{wang2023ret,xia2024text2loc,xia2026text2locpp,liu2025mncl,shang2025mambaplace,xu2025cmmloc}. These works mainly answer whether candidates are relevant and whether predictions are accurate. After candidates are produced, however, an independent question remains: when several candidates each provide a coordinate, which one should be output?

Region relevance does not guarantee a correct coordinate. A semantically relevant region may yield a wrong coordinate because of inaccurate local correspondence or offset estimation, whereas a lower-ranked region may contain an accurate prediction. When candidate cells overlap or are adjacent, cell identity and metric localization success are especially easy to separate: a ground-truth (GT) cell can contain a prediction several meters away, and a neighboring cell can produce a coordinate within the threshold. If the final output is still determined by retrieval order, valid results already produced by the fine-localization stage may never be selected. We therefore evaluate the pair ``candidate cell and its existing Fine coordinate'' as a complete localization hypothesis.

\begin{figure*}[!t]
\centering
\includegraphics[width=0.95\textwidth]{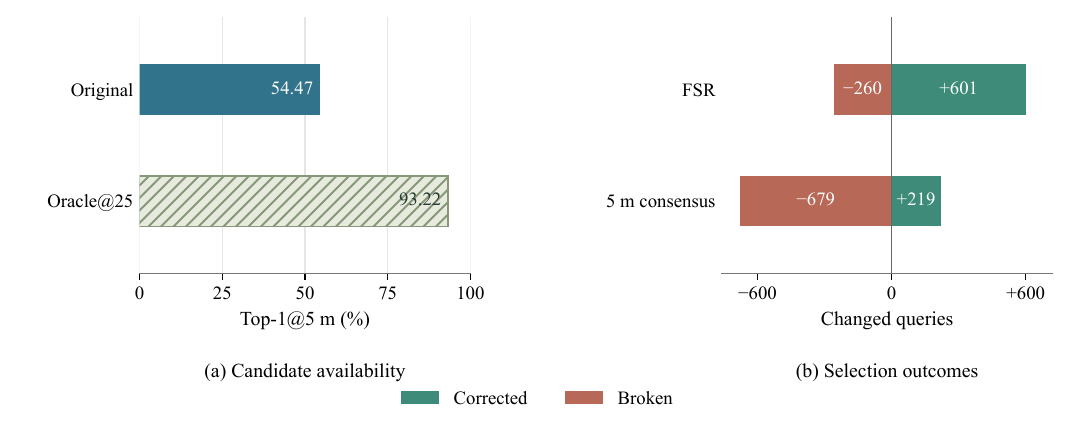}
\caption{Candidate availability and selection on MNCL$\times$KITTI VAL with a fixed Top-25 prediction pool. (a) Original and Oracle@25 Top-1@5\,m. (b) Queries corrected and broken by two training-free rankings relative to Original. The oracle uses GT only to measure recoverable coverage. Query counts, net gains, and the paired interval are discussed in the text.}
\label{fig:pool}
\end{figure*}

This problem is directly observable within a fixed prediction pool (Fig.~\ref{fig:pool}). On MNCL$\times$KITTI VAL, 3,187 queries from one scene form 1,116 pose clusters. Original-order Top-1@5\,m is 54.47\%, whereas Oracle@25 reaches 93.22\% using the same Fine predictions. Among the 1,451 Original failures, 1,235 already contain a successful candidate (85.11\%); the remaining 216 have no successful prediction in the pool. The 38.75\,pp headroom separates candidate availability from final selection. Its use requires a ranking rule that balances recovered failures against newly broken successes: FSR corrects 601 and breaks 260 queries, yielding a net gain of 10.70\,pp (paired 95\% interval $[8.68,12.63]$), whereas 5\,m position consensus corrects 219 and breaks 679. The interval uses 2,000 paired pose-cluster bootstrap resamples (seed 123). These are development-set diagnostics; Oracle@25 uses GT for evaluation rather than inference. Complete results are in Supplement S5.1.

SCAR constructs prediction-point-conditioned spatial evidence for this decision stage. Its core idea is to interpret a directional relation in a query as a function of position: ``east of an object'' has a verifiable truth value only once a position to be tested is specified. Consequently, ``the region contains a position supporting the query'' and ``the backbone's actual predicted point is supported'' answer two different questions: whether the region can be correct, and whether the coordinate about to be output is compatible. SCAR reads spatial support at the actual Fine prediction point (Fine-conditioned Evidence Readout), retains center evidence, ranks candidates with a scorer, and returns the original coordinate of the first hypothesis.

This design must address three interrelated difficulties. First, color and category matching can only confirm semantic presence; multiple directional constraints must hold simultaneously at the same map position. Second, a region center or a support peak inside the region cannot substitute for checking the actual Fine coordinate. Third, retrieval similarity, object presence, spatial support, and geometric offsets have different scales and failure modes, so a single handcrafted score is unlikely to adapt to all backbones and domains. SCAR integrates these complementary signals with a shared scorer, constructs supervision from actual localization success, and selects the training state by validation localization metrics. Candidates and Fine coordinates remain unchanged, so performance changes can be interpreted within the same hypothesis set as differences in selection.

The contributions are as follows.
\begin{enumerate}
\item We formulate the final output of coarse-to-fine localization as a selection problem over fixed localization hypotheses: a candidate cell and its existing Fine coordinate are evaluated jointly, explicitly separating region retrieval correctness from metric localization success.
\item We propose Fine-conditioned Evidence Readout, which reads directional-relation support at the backbone's actual predicted point and organizes it with center, semantic, retrieval, and geometric evidence into a 23-D candidate representation; a candidate-level listwise ranker is trained with localization-success labels.
\item We evaluate fixed-pool reranking on five domains and test the Fine readout and the complete scoring scheme with two controlled VAL studies. On corrected Des4Pos$\times$KITTI, the Fine readout improves Top-1@5\,m by 5.77 and 5.99\,pp over a center surrogate and over removing the support pathway, respectively. On MambaPlace$\times$nuScenesFull, the complete scoring scheme improves it by 13.98\,pp over FSR.
\end{enumerate}

Section~\ref{sec:related} reviews related work, Section~\ref{sec:method} defines the method and training objective, Section~\ref{sec:setup} presents the experimental protocol, Sections~\ref{sec:results}--\ref{sec:efficiency} analyze localization results, the spatial-evidence mechanism, and efficiency, and Section~\ref{sec:discussion} discusses applicability and limitations. Model-selection rules, provenance, and supplementary readings are provided in the supplement.

\section{Related Work}\label{sec:related}
\subsection{Cross-modal Localization Between Language and 3-D Maps}
Place recognition usually maps a local observation to a retrievable global descriptor. NetVLAD constructs image place representations through learnable aggregation~\cite{arandjelovic2016netvlad}, and PointNetVLAD extends this idea to point-cloud retrieval~\cite{uy2018pointnetvlad}. Language queries add cross-modal semantic correspondence and relational constraints: the model must determine which map objects are relevant from a local description and infer the position they jointly constrain. Visual place recognition has further improved descriptors through multi-scale local--global fusion~\cite{hausler2021patchnetvlad}, universal place representations~\cite{keetha2024anyloc}, viewpoint-robust training~\cite{berton2023eigenplaces}, and global feature mixing~\cite{alibey2023mixvpr}. 3-D referring-expression localization such as ScanRefer outputs a target object box~\cite{chen2020scanrefer}; the text-to-map-position localization considered here outputs the describer or query position. In 3-D visual grounding, 3DVG-Transformer~\cite{zhao2021_3dvg}, TransRefer3D~\cite{he2021transrefer3d}, and 3D-SPS~\cite{luo2022_3dsps} strengthen language--scene correspondence through relation modeling, entity--relation attention, and progressive key-point sampling, respectively; ReferIt3D~\cite{achlioptas2020referit3d} provides a fine-grained referring evaluation protocol.

Text2Pos divides language localization into map-cell retrieval, object matching, and position estimation~\cite{kolmet2022text2pos}. RET explicitly enhances relation modeling in text and object encoders and refines positions through fine-grained cross-modal matching~\cite{wang2023ret}. Text2Loc combines global contrastive learning with Fine prediction that does not require explicit instance matching~\cite{xia2024text2loc}; Text2Loc++ further studies complex language, masked-instance training, modality-aware contrastive learning, and cascaded cross-attention fine localization~\cite{xia2026text2locpp}. These methods improve candidate quality from both representation and estimation; SCAR evaluates the final output hypothesis after they have produced candidate coordinates.

MNCL organizes negative constraints at global, instance, and relation levels to sharpen the distinction between text and mismatched places~\cite{liu2025mncl}; MambaPlace uses attention-Mamba mechanisms for sequence interaction between text and point clouds~\cite{shang2025mambaplace}; CMMLoc addresses descriptions that cover only part of the environment with a Cauchy-mixture prior and direction fusion~\cite{xu2025cmmloc}. Des4Pos employs multi-scale point-cloud representation, staged text encoding, and residual attention for place recognition and offset prediction~\cite{shang2025des4pos}. Geometric encoders such as PointNet++ and pretrained language models such as T5 provide reusable representation components for these backbones~\cite{qi2017pointnetpp,raffel2020t5}.

\subsection{Spatial Relations and Position-Conditioned Evidence}
Directional relations in language depend on the reference position and cannot be treated as fixed object attributes. In image localization, SuperGlue~\cite{sarlin2020superglue} and LoFTR~\cite{sun2021loftr} establish correspondences through local feature matching and can serve as a basis for geometric verification. RET and CMMLoc already incorporate relation or direction information into cross-modal localization~\cite{wang2023ret,xu2025cmmloc}. PosEviLoc constructs a spatial-evidence field conditioned on a hypothesized query position and performs place retrieval with a compact representation, a lightweight MLP, and listwise supervision, supporting plug-in reranking of external candidates~\cite{shang2026poseviloc}. That work shows that position-conditioned evidence can be a retrieval signal independent of backbone prediction.

SCAR differs from PosEviLoc in readout position and supervision target. PosEviLoc constructs evidence for a hypothesized query position and targets place retrieval; SCAR reads directional support at the candidate's existing Fine prediction point, constructs positives from the final localization success of that point, and outputs the original coordinate. Strong support somewhere in a region does not guarantee that the region's Fine prediction lies at a supported position; verification is therefore moved from the region or cell center to the coordinate that will be output.

\subsection{Hypothesis Verification and Listwise Learning}
Verification after candidate generation is a general decomposition in localization. InLoc evaluates candidate poses by view synthesis, illustrating a design in which prediction and verification use complementary information~\cite{taira2018inloc}. SCAR addresses hypothesis selection for language-to-2-D-map positions, using object semantics and directional compatibility as verification evidence.

Learning to rank treats candidates in the same list as a competing set, and listwise cross-entropy is a mature objective~\cite{cao2007listwise}. Pairwise RankNet~\cite{burges2005ranknet} and LambdaRank, which directly optimizes ranking metrics~\cite{burges2006lambdarank}, trace the progression from pairwise to listwise objectives; DLCM~\cite{ai2018dlcm} and SetRank~\cite{pang2020setrank} further model candidate competition within list context. In retrieval, ColBERT late interaction~\cite{khattab2020colbert} and k-reciprocal reranking~\cite{zhong2017kreciprocal} provide fine-grained query--candidate matching and neighborhood-relation reranking, respectively. SCAR adopts a listwise objective, uses a uniform distribution over localization-successful candidates as supervision, and selects the training state by validation Top-1@5\,m. Positives are defined by the metric error of the candidate's Fine coordinate and scene consistency.

\section{Method}\label{sec:method}
\begin{figure*}[!t]
\centering
\includegraphics[width=1\textwidth]{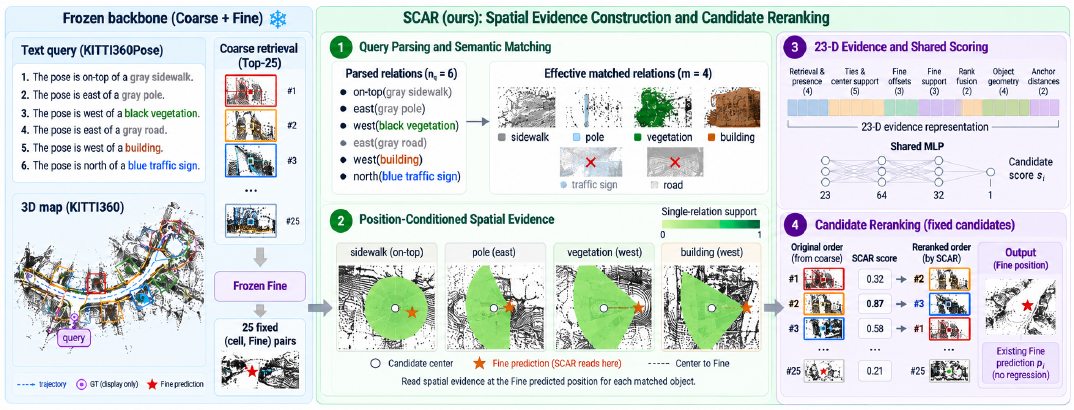}
\caption{SCAR overview. A frozen Coarse--Fine backbone supplies 25 cell--position hypotheses. Query relations are matched to map objects, spatial support is read at each existing Fine prediction, and the resulting 23-D representation is normalized with TRAIN statistics and scored by a shared MLP. SCAR reranks the fixed hypotheses and returns the existing Fine coordinate of the selected candidate. Point-cloud layouts, relations, support fields, and scores are illustrative.}
\label{fig:roadmap}
\end{figure*}

SCAR organizes candidates produced by a frozen backbone into localization hypotheses. Fig.~\ref{fig:roadmap} groups the procedure into query parsing and semantic matching, position-conditioned support readout, shared scoring, and candidate reranking. A direction is evaluated relative to the tested position rather than stored as an intrinsic object attribute. The effective relation count $m$ records query relations admitted to directional scoring, not the number already satisfied at that point. The Fine-support features are evaluated at the existing predicted coordinate, with center support retained as complementary evidence. TRAIN-fitted normalization and a shared $23\!\rightarrow\!64\!\rightarrow\!32\!\rightarrow\!1$ scorer produce within-list scores; descending score order selects an existing Fine prediction without coordinate regression.

Given a frozen candidate hypothesis set, each item retains both a candidate map cell and the Fine world coordinate already predicted for that candidate. SCAR learns a permutation of this set and outputs the existing coordinate of the first hypothesis. During training, only scorer parameters are updated; localization labels and validation selection are defined by \eqref{eq:label} and \eqref{eq:select}, respectively. Backbone predictions and candidate sets remain fixed during scorer training and evaluation.

To separate spatial verification from learned fusion, we
define three training-free controls. Semantic Matching is a training-free control constructed in this study. It augments retrieval similarity with color--category matching coverage of query mentions, without evaluating directional support at the Fine position. Rank Fusion combines the canonical candidate rank with the Fine-support rank into one ordering; FSR (Fine-position Support Ranking) directly sorts by support at the predicted point. All three share the same candidates and coordinates as SCAR and measure the increment of the complete learned scheme.

\subsection{Problem Definition and Frozen Interface}
Given a query $q$ and a gallery of $N$ map cells, Coarse retrieval produces $K$ candidates and an original pool order. The cosine similarity $g_i$ between the text and the candidate's global representation is also recorded as SCAR's \texttt{cos} feature. Fine produces a normalized 2-D prediction $u_i$ for candidate $c_i$. For the frozen interface with square normalized cells, the world coordinate is
\begin{equation}
p_i = o_i + L_i u_i,
\label{eq:coord}
\end{equation}
where $o_i$ is the 2-D origin of the cell boundary and $L_i$ is the cell scale in the frozen data definition. Each dataset uses its own frozen coordinate adaptation and cell scale.

SCAR receives the fixed hypothesis set
\begin{equation}
\mathcal{H}_K(q)=\{(c_i,g_i,u_i,p_i)\}_{i=1}^{K},\qquad K=25,
\label{eq:hyp}
\end{equation}
and outputs a permutation $\pi$; the final position is $p_{\pi(1)}$.

\subsection{Query Parsing and Semantic Presence}
A rule-based parser uses each frozen dataset's vocabulary and description templates to represent a query as a sequence of relations
\begin{equation}
R(q)=(r_1,\ldots,r_{n_q}),\qquad r_j=(a_j,\ell_j,d_j),
\label{eq:query}
\end{equation}
where $a_j$, $\ell_j$, and $d_j$ are the color, object category, and direction of the query position relative to the object. Repeated descriptions are kept as separate slots. Directions are interpreted with respect to the position under test.

Let $t=(a,\ell)$ be a color--category combination, $n_q(t)$ its occurrence count in the query, and $n_c(t)$ its object-inventory count in the candidate. Semantic presence is defined as
\begin{equation}
\mathrm{sp}(q,c)=\frac{1}{n_q}\sum_{t\in T(q)}\min\{n_q(t),\,n_c(t)\}.
\label{eq:presence}
\end{equation}
This count prevents one object from covering repeated descriptions arbitrarily many times, but it does not check direction. Object categories and colors come from the map-object attributes used by the backbone; the query's true position and GT-matched identity are used only to construct training labels and for evaluation.

\subsection{Position-Conditioned Directional Support Field}
A $33\times33$ grid $G_c$ is established in the candidate's normalized 2-D region. For a grid point and an object point set, the relative displacement to the nearest 2-D point determines east, west, north, or south, or the on-top convention when the 2-D distance is below $0.05$ cell units. Non-on-top directions are determined by the dominant coordinate axis and its sign.

The current direction field first takes the first 40 objects in storage order and then skips empty point sets. Let $n_c(t)$ be the number of objects of color--category combination $t$ among them. In the audited frozen implementations of corrected Des4Pos$\times$KITTI and MambaPlace$\times$nuScenesFull, only groups whose object count covers all query slots enter directional scoring.

Let $E(q,c)=\{t:n_c(t)\geq n_q(t)\}$ be the set of description groups for which the candidate has at least as many objects as the query requires, and let
\begin{equation}
m(q,c)=\sum_{t\in E(q,c)} n_q(t)
\label{eq:effective}
\end{equation}
be the total number of query descriptions in these admitted groups. For each available group, a one-to-one assignment between description slots and objects maximizes the number of directional relations that hold simultaneously at the grid point. The maximum satisfied counts of all groups are summed to obtain the support field $h(q,c,u)=H(q,c,g_c(u))$ at position $u$; when a group has too few objects, the whole group is skipped, consistent with the definition of \eqref{eq:effective}.

For a position $u$, $h$ is read at the nearest grid point $g_c(u)$. We define
\begin{align}
S_{\mathrm{match}}(q,c,u)&=
\begin{cases}
h(q,c,u)/m(q,c), & m(q,c)\geq 4,\\
0, & m(q,c)<4,
\end{cases}\label{eq:smatch}\\
S_{\mathrm{query}}(q,c,u)&=
\begin{cases}
h(q,c,u)/n_q, & m(q,c)\geq 4,\\
0, & m(q,c)<4.
\end{cases}\label{eq:squery}
\end{align}
A fixed threshold of four controls whether directional scoring is enabled. Support is approximated by reading the nearest grid point to the Fine prediction; for out-of-region predictions, the nearest boundary grid point serves as a support proxy while the original predicted coordinate is retained. When $m<4$, directional support is set to zero and the remaining evidence still participates in scoring.

With the cell center $u_0=(0.5,0.5)$, we construct
\begin{equation}
\begin{aligned}
\texttt{sc\_center}_i&=S_{\mathrm{match}}(q,c_i,u_0),\\
\texttt{sc\_fine}_i&=S_{\mathrm{match}}(q,c_i,u_i),\\
\texttt{sat\_fine}_i&=S_{\mathrm{query}}(q,c_i,u_i).
\end{aligned}
\label{eq:readouts}
\end{equation}
The three readouts have different meanings. \texttt{sc\_center} is the support at the cell center; \texttt{sc\_fine} is the support at the actual Fine prediction point. Both are point samples of the same support field and neither takes the maximum support in the region. SCAR's Fine-conditioned Evidence Readout uses \texttt{sc\_fine}, converting ``somewhere in the region is supported'' into ``the coordinate about to be output is supported.'' \texttt{sat\_fine} shares the satisfied count with \texttt{sc\_fine} but retains all query descriptions in the denominator, measuring support relative to the complete query. If a query contains six descriptions, four enter directional scoring, and three hold at the Fine point, the two values are $3/4$ and $3/6$. Sorting only by \texttt{sat\_fine} without learning defines FSR; it uses the same readout as SCAR but does not learn joint weights. In the implementation, \texttt{sc\_center} and \texttt{sc\_fine} both use $m(q,c)$ as the denominator, whereas \texttt{sat\_fine} uses $n_q$; the three are not interchangeable.

\subsection{Complementary Evidence and the 23-D Representation}
First, candidates are sorted stably in descending order of $g_i+0.5\,\mathrm{sp}(q,c_i)$ to obtain $P_0$ (Semantic Matching). When the maximum presence is attained by several candidates exactly tied, only those candidates are stably reordered by \texttt{sc\_center} within their original slots to obtain $P_2$; all other slots are unchanged. The gate is triggered by exact ties in the presence score.

Semantic Matching outputs $P_0$, the stable descending order of the retrieval score plus half the presence score; exact ties retain the input candidate order. $P_2$ only resolves the tie block at maximum presence and provides a canonical order for subsequent features and rank fusion; it does not change Semantic Matching's output.

For MNCL, the original Top-25 pool and its order are retained from the frozen multi-level normalized composite score; SCAR's \texttt{cos} feature is computed from a separately L2-normalized global-layer representation. Using $P_2$ as the canonical order, let $r_{P_2}(i)$ and $r_f(i)$ be the candidate's zero-based $P_2$ rank and \texttt{sc\_fine} descending rank. The Rank Fusion (Borda-style) score is $[r_{P_2}(i)+r_f(i)]/[2(K-1)]$, output in ascending order; exact ties in \texttt{sc\_fine} or fusion scores retain $P_2$ order. FSR sorts by the support value of \eqref{eq:readouts} in descending order and also retains $P_2$ order on exact ties. The anchor is the first item of canonical $P_2$.

Seven evidence groups form the 23-D representation, corresponding to judgments from region to coordinate. Coarse ranking and presence (4-D) record the candidate's prior relevance and how well query objects can be satisfied; candidate organization and center evidence (5-D) record list position, presence tie structure, and directional support at the region center; Fine position relative to the center (3-D) characterizes the direction and magnitude of the prediction's displacement from the center; Fine spatial support (3-D) gives the satisfied proportion and support level of directional relations at the predicted point; rank fusion (2-D) combines canonical rank and Fine-support rank into a within-list relative position; object and neighborhood geometry (4-D) describes the number and distances of matchable objects near the prediction; relative anchor position (2-D) records the displacement of the prediction and center from the list anchor. The seven groups contain 4, 5, 3, 3, 2, 4, and 2 dimensions, totaling 23. The canonical column names, 0-based indices, and code mapping are listed in Supplement S1.

Retrieval and presence answer ``whether the region may be relevant''; center support answers ``how compatible the center is with the query relations''; Fine support answers ``whether the backbone's actual output coordinate is compatible''; offsets and object geometry answer ``what local structure surrounds the prediction''; rank fusion and anchor distance provide relative references within the list. A single support score may overestimate a displaced prediction because a supported position exists somewhere in the region, or underestimate a valid candidate whose Fine point falls at the edge of the support field; only by placing region-level and point-level evidence side by side can the scorer distinguish these cases.

Among the offset features are the 2-D displacement of the normalized Fine position from the center and its magnitude. The object count is the number of matched objects recorded by the feature-construction path and differs from the number $m$ of effective relations entering directional scoring in \eqref{eq:effective}: the former counts the object inventory, whereas the latter counts query slots eligible for directional scoring, and the two can disagree when repeated descriptions are under-covered.

In the two configurations that have completed code-level auditing---corrected Des4Pos$\times$KITTI and MambaPlace$\times$nuScenesFull---columns 19 and 20 under zero-based indexing (\texttt{dist\_nearest\_matched}, \texttt{dist\_centroid\_matched}) are fixed constants of $3.0$ during feature construction, become identically zero after scorer normalization, and therefore do not change the scorer input. This conclusion covers only these two configurations. The actual values of these two columns were not audited item by item in the remaining configurations; it cannot be inferred that they behave identically, nor can the observation be extended to ``the two columns have no effect on results.''

\subsection{Shared Scoring with Localization-Success Supervision}
Each candidate feature $x_i$ is normalized by the mean and standard deviation fitted only on valid TRAIN lists, with near-constant columns handled by constants recorded in the frozen head, yielding $z_i$. The shared scorer is
\begin{equation}
s_i=f_\theta(z_i),\qquad f_\theta:23\rightarrow64\rightarrow32\rightarrow1,
\label{eq:scorer}
\end{equation}
with ReLU hidden layers. The three-layer biased fully connected scorer has 3,649 parameters and shares them across all candidates, evaluating every hypothesis by the same rule. Scores are used to compare candidates within the same list.

The supervision target depends on the coordinate that must ultimately be returned, not merely on candidate-cell identity. Neighboring or overlapping regions may produce predictions that satisfy the metric threshold, and a GT cell may also contain an incorrect Fine prediction. We therefore test each candidate's output coordinate directly and define the localization-success label
\begin{equation}
y_i=\mathbb{1}\big[\mathrm{scene}(c_i)=\mathrm{scene}(q)\big]\cdot
\mathbb{1}\big[\|p_i-p_q^{*}\|_2\leq5\,\mathrm{m}\big],
\label{eq:label}
\end{equation}
where $p_q^{*}$ is the GT query position.

For a training list $i=1,\ldots,K$, define the target distribution $t_i=y_i/\sum_j y_j$ and minimize
\begin{equation}
\mathcal{L}(q)=-\sum_{i=1}^{K}t_i\log\frac{\exp(s_i)}{\sum_{j=1}^{K}\exp(s_j)}.
\label{eq:loss}
\end{equation}
Lists with no positives or with all positives are excluded from head training and normalization fitting under the current protocol; VAL and TEST retain all queries. The backbone remains frozen. Training uses AdamW with learning rate $10^{-3}$, weight decay $10^{-4}$, list batch size 512, and 30 epochs. With a fixed backbone, scorer initialization seed 0 is primary, and seeds 1 and 2 are used for sensitivity analysis.

\subsection{Validation Selection Consistent with the Localization Objective}
Each epoch computes Top-1@5\,m and Cell R@1 on the complete validation query set. The formal results use a localization-metric-first selection rule: take the state with the highest Top-1@5\,m; break exact ties by Cell R@1; break remaining ties by the earlier epoch:
\begin{equation}
e_{\mathrm{SCAR}}=\arg\max^{\mathrm{lex}}_{e}\big(\ell_e,\,a_e,\,-e\big),
\label{eq:select}
\end{equation}
where $\ell_e$ and $a_e$ are the epoch's validation Top-1@5\,m and Cell R@1. The last lexicographic term guarantees that the earlier epoch is chosen under a strict tie. This rule is fixed across all configurations; the rule or the seed is never switched according to TEST readings.

\subsection{Inference and Computational Boundary}
Algorithm~\ref{alg:scar} summarizes inference. Because the output is a permutation of the same hypothesis set, all compared methods have identical Cell R@25 and Top-25@5\,m coverage. A successful hypothesis missed by coarse retrieval, and a query for which every Fine prediction fails, cannot be corrected by reranking alone.

\begin{algorithm}[!t]
\caption{Spatial-consistency-aware reranking over a fixed hypothesis set.}
\label{alg:scar}
\begin{algorithmic}[1]
\STATE Parse the query; compute semantic presence and the directional support field for every candidate.
\STATE Read spatial support at the cell center and at the candidate's actual Fine prediction point.
\STATE Form \texttt{sc\_center}, \texttt{sc\_fine}, \texttt{sat\_fine}, canonical ranks, and the anchor; build the 23-D feature vector in the fixed column order.
\STATE Normalize with TRAIN statistics stored in the checkpoint selected by \eqref{eq:select}; score all candidates with the shared MLP.
\STATE Sort by descending score; retain $P_2$ order on exact ties.
\STATE Output the existing Fine world coordinate of the first-ranked candidate.
\end{algorithmic}
\end{algorithm}

\section{Experimental Setup}\label{sec:setup}
\subsection{Backbones and Derived Localization Benchmarks}
Experiments cover target-domain adaptations of four public backbones---Text2Loc, Text2Loc++, MNCL, and MambaPlace---and separately report a Des4Pos reimplementation reference~\cite{xia2024text2loc,xia2026text2locpp,liu2025mncl,shang2025mambaplace,shang2025des4pos}. The five localization benchmarks are KITTI360Pose, PandaSetPose, nuScenesPose, S3DISPose, and ParisLillePose/NPM3D. KITTI360Pose comes from Text2Pos's language-localization construction on KITTI-360~\cite{kolmet2022text2pos,liao2022kitti360}; the other benchmarks use frozen derived localization packages built on PandaSet, nuScenes, S3DIS, and Paris-Lille-3D~\cite{xiao2021pandaset,caesar2020nuscenes,armeni2016s3dis,roynard2018parislille}. Table~\ref{tab:splits} lists the splits and sample sizes.

Each configuration uses its own frozen Coarse/Fine pipeline and domain adaptation. The 20 public-backbone entries are results after target-domain adaptation, and comparisons are made within each configuration's fixed candidate pool.

The input semantics of the derived data must be read together with their names. Object representations for PandaSet and nuScenes consist of surface samples from annotated boxes and object-level colors, unlike raw per-point LiDAR; nuScenes retains 2--6 original descriptions, and comparisons use only the full-query version. Fine object budgets differ across backbones on S3DIS. GT cells for PandaSet and nuScenes are defined by the nearest retained anchor, so some positions may fall outside a cell; evaluation keeps the original annotations. Raw data come from KITTI-360~\cite{liao2022kitti360}, PandaSet~\cite{xiao2021pandaset}, nuScenes~\cite{caesar2020nuscenes}, S3DIS~\cite{armeni2016s3dis}, and Paris-Lille-3D~\cite{roynard2018parislille}.

\begin{table}[!t]
\centering
\caption{Frozen localization splits. Counts are localization requests; one request may contain several descriptions.}
\label{tab:splits}
\scriptsize
\setlength{\tabcolsep}{2.5pt}
\begin{tabular}{@{}lrrrp{3.15cm}@{}}
\toprule
Benchmark & TRAIN & VAL & TEST & Split-unit definition \\
\midrule
KITTI360Pose & 28,807 & 3,187 & 11,505 & 30 m cells; six descriptions; TEST covers three scenes \\
PandaSetPose & 4,000 & 2,000 & 2,000 & six descriptions per pose; nearest retained anchor cell \\
nuScenesPose Full & 15,113 & 7,635 & 7,403 & 2--6 original descriptions; 210 TEST scenes with queries \\
S3DISPose & 1,084 & 544 & 542 & rooms as split unit; four descriptions; TEST 68 rooms \\
ParisLillePose/NPM3D & 120 & 128 & 116 & 20 m cells; three descriptions; TEST 29 query scenes \\
\bottomrule
\end{tabular}
\end{table}

\subsection{Unified Candidate Budget and Control Methods}
Each configuration fixes 25 candidates and uses same-source Fine predictions for all of them. Original Ranking denotes the backbone's own candidate order and its Fine outputs; Semantic Matching, Rank Fusion, and FSR are the training-free controls defined in Section~\ref{sec:method}; SCAR is the learned method with localization-metric-first model selection.

The measured native pipelines of Text2Loc, Text2Loc++, MNCL, and MambaPlace use the Top-10 Fine candidates. Same-pool accuracy comparisons use a unified Top-25 candidate budget. Efficiency experiments separately retain the native Top-10 and the cost of extending to Top-25, so that the Fine forward passes for the additional 15 candidates are not hidden inside reranking overhead.

The scorer for each target domain is fitted on that domain's TRAIN split. Main tables fix backbone training seed 0 and scorer initialization seed 0; additional backbone repeats and scorer seeds are reported separately.

\subsection{Metrics and Paired Uncertainty}
Cell R@$k$ checks whether the first $k$ candidates contain the GT cell. Top-$k$@$r$ checks whether the Fine predictions of the first $k$ candidates contain a successful item in the same scene within $r$ meters of the GT. The primary metric is Top-1@5\,m, and Cell R@1 is also reported. All queries enter the denominator; queries without a successful position in the candidate pool still count as failures.

Paired analyses fix queries, candidates, and Fine predictions. They bootstrap the per-query hit difference of the two rankings 2,000 times with random seed 123, over the pose, room, or scene clusters registered by each frozen protocol. Different cluster definitions are labeled separately. KITTI has only three TEST scenes, so a scene-cluster interval describes variation conditional on these three scenes; a single-scene VAL does not support between-scene inference.

\begin{figure*}[!t]
\centering
\includegraphics[width=0.95\textwidth]{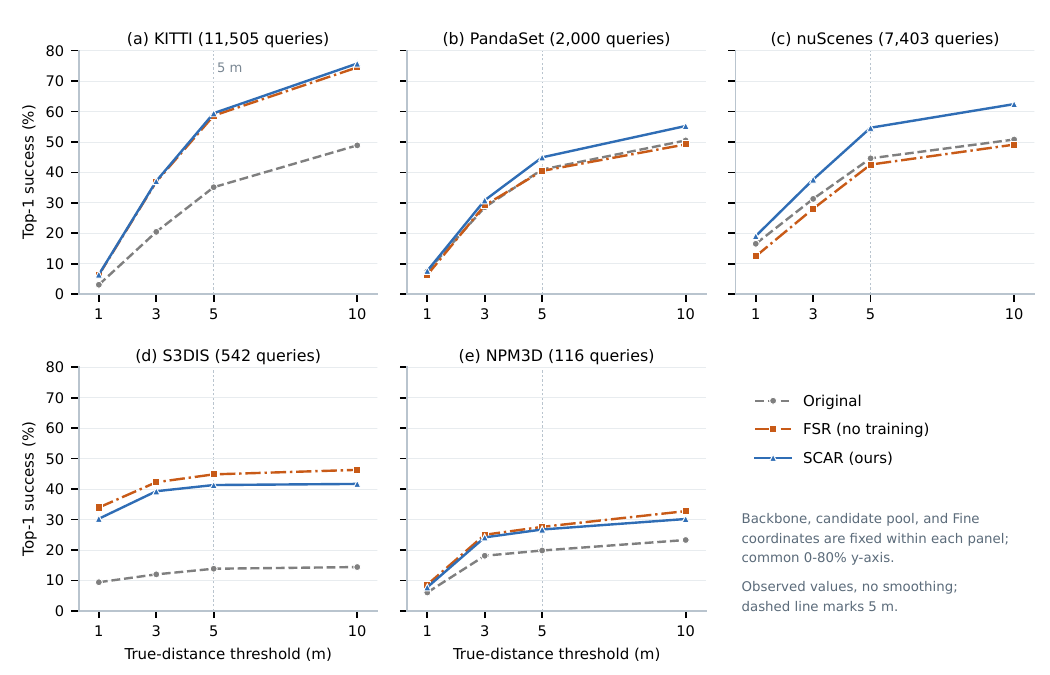}
\caption{TEST Top-1 localization success as a function of the true-distance threshold (1, 3, 5, and 10\,m). All five panels use the Text2Loc backbone and its fixed Top-25 candidate pool and Fine coordinates on KITTI, PandaSet, nuScenes, S3DIS, and NPM3D. Original, FSR, and SCAR are three rankings of the same candidate set. The vertical axis is 0--80\% in every panel; the gray dashed line marks the 5\,m primary metric. Curves connect observed values without smoothing. All TEST comparisons are exploratory; Table~\ref{tab:top1} reports the four-threshold SCAR results, and Supplement Table S4 gives additional comparisons.}
\label{fig:thresholds}
\end{figure*}

The TEST splits have participated in development and analysis, so aggregate comparisons are exploratory evidence. Intervals follow each frozen computation record, the statistical unit is labeled with the result, and no multiple-comparison correction is applied. Confidence levels or interval-construction details not fully registered in the source are listed in Supplement S3. Table~\ref{tab:top1} reports SCAR at all four distance thresholds. Fig.~\ref{fig:thresholds} shows the corresponding curves for Text2Loc.

\section{Localization and Retrieval Results}\label{sec:results}
\subsection{Localization Within a Fixed Hypothesis Pool}
Tables~\ref{tab:top1} and~\ref{tab:cell} retain 25 configurations with backbone training seed 0 and scorer initialization seed 0. Twenty are target-domain adaptations of the four public backbones, and five are the Des4Pos reimplementation; all five domains use corrected same-pool TEST results. Table~\ref{tab:top1} reports Top-1 localization success at true-distance thresholds of 1, 3, 5, and 10\,m, and Table~\ref{tab:cell} reports same-pool Cell R@1, R@3, R@5, and R@10. Rows identify the localization backbone. The rankings in each comparison use the same backbone's candidates and Fine predictions. In both tables, the parenthesized difference uses Original as the baseline, i.e., SCAR minus Original for the same backbone and dataset; percentages keep two decimals, and differences are computed from unrounded source values and then rounded to two decimals. MambaPlace$\times$PandaSet uses a new ep2 trajectory whose protocol alignment with the other configurations is unresolved; that row is a reference only. Table~\ref{tab:fullcontrols} compares SCAR with Semantic Matching, Rank Fusion, and FSR at Top-1@5\,m and Cell R@1. Supplement Tables S4 and S5 extend this comparison to the other distance thresholds and cell ranks.

\begin{table*}[!t]
\centering
\caption{TEST localization success (\%) of SCAR at four distance thresholds. Each cell shows the success rate above and the difference from Original (pp) below. The primary threshold is 5\,m.}
\label{tab:top1}
\small
\setlength{\tabcolsep}{2pt}
\renewcommand{\arraystretch}{1.12}
\begingroup
\renewcommand{\SCARresult}[2]{\shortstack[c]{#1\\{\footnotesize #2}}}
\begin{tabular*}{\textwidth}{@{\extracolsep{\fill}}lcccccccccccc@{}}
\toprule
 & \multicolumn{4}{c}{\textbf{KITTI}} & \multicolumn{4}{c}{\textbf{S3DIS}} & \multicolumn{4}{c}{\textbf{PandaSet}} \\
\cmidrule(lr){2-5} \cmidrule(lr){6-9} \cmidrule(lr){10-13}
Backbone & 1\,m & 3\,m & \textbf{5\,m} & 10\,m & 1\,m & 3\,m & \textbf{5\,m} & 10\,m & 1\,m & 3\,m & \textbf{5\,m} & 10\,m \\
\midrule
Text2Loc & \SCARresult{6.46}{\textbf{(+3.36)}} & \SCARresult{37.16}{\textbf{(+16.68)}} & \SCARresult{59.45}{\textbf{(+24.31)}} & \SCARresult{75.77}{\textbf{(+26.88)}} & \SCARresult{30.26}{\textbf{(+20.85)}} & \SCARresult{39.30}{\textbf{(+27.31)}} & \SCARresult{41.33}{\textbf{(+27.49)}} & \SCARresult{41.70}{\textbf{(+27.31)}} & \SCARresult{7.70}{\textbf{(+0.15)}} & \SCARresult{30.85}{\textbf{(+2.30)}} & \SCARresult{44.95}{\textbf{(+4.00)}} & \SCARresult{55.25}{\textbf{(+4.75)}} \\
Text2Loc++ & \SCARresult{8.29}{\textbf{(+3.78)}} & \SCARresult{41.23}{\textbf{(+15.09)}} & \SCARresult{64.15}{\textbf{(+20.43)}} & \SCARresult{78.86}{\textbf{(+21.10)}} & \SCARresult{34.50}{\textbf{(+18.27)}} & \SCARresult{44.10}{\textbf{(+24.17)}} & \SCARresult{45.94}{\textbf{(+25.28)}} & \SCARresult{47.97}{\textbf{(+26.20)}} & \SCARresult{4.55}{\textbf{(+0.35)}} & \SCARresult{29.45}{\textbf{(+4.00)}} & \SCARresult{44.30}{\textbf{(+6.95)}} & \SCARresult{56.30}{\textbf{(+8.85)}} \\
MNCL & \SCARresult{7.20}{\textbf{(+2.83)}} & \SCARresult{40.40}{\textbf{(+11.26)}} & \SCARresult{63.82}{\textbf{(+14.23)}} & \SCARresult{79.69}{\textbf{(+13.99)}} & \SCARresult{22.88}{\textbf{(+18.08)}} & \SCARresult{36.90}{\textbf{(+29.70)}} & \SCARresult{38.38}{\textbf{(+30.44)}} & \SCARresult{39.11}{\textbf{(+31.18)}} & \SCARresult{4.65}{(-0.30)} & \SCARresult{27.20}{(-2.10)} & \SCARresult{42.05}{(-0.15)} & \SCARresult{52.50}{(-0.10)} \\
MambaPlace$^{\dagger}$ & \SCARresult{7.21}{\textbf{(+3.95)}} & \SCARresult{36.01}{\textbf{(+17.49)}} & \SCARresult{57.07}{\textbf{(+25.09)}} & \SCARresult{72.92}{\textbf{(+28.34)}} & \SCARresult{22.14}{\textbf{(+16.24)}} & \SCARresult{38.19}{\textbf{(+27.68)}} & \SCARresult{40.77}{\textbf{(+28.78)}} & \SCARresult{42.07}{\textbf{(+28.60)}} & \SCARresult{1.80}{\textbf{(+0.45)}} & \SCARresult{14.35}{\textbf{(+5.65)}} & \SCARresult{22.15}{\textbf{(+8.25)}} & \SCARresult{34.65}{\textbf{(+11.85)}} \\
Des4Pos (reimpl.) & \SCARresult{6.62}{\textbf{(+4.17)}} & \SCARresult{36.11}{\textbf{(+19.06)}} & \SCARresult{57.84}{\textbf{(+28.47)}} & \SCARresult{73.53}{\textbf{(+32.37)}} & \SCARresult{22.32}{\textbf{(+14.21)}} & \SCARresult{36.35}{\textbf{(+22.51)}} & \SCARresult{37.82}{\textbf{(+23.25)}} & \SCARresult{39.11}{\textbf{(+23.25)}} & \SCARresult{3.25}{\textbf{(+0.60)}} & \SCARresult{21.60}{\textbf{(+8.95)}} & \SCARresult{31.55}{\textbf{(+11.20)}} & \SCARresult{44.65}{\textbf{(+14.65)}} \\
\bottomrule
\end{tabular*}
\par\medskip
\begin{tabular*}{\textwidth}{@{\extracolsep{\fill}}lcccccccc@{}}
\toprule
 & \multicolumn{4}{c}{\textbf{nuScenes}} & \multicolumn{4}{c}{\textbf{NPM3D}} \\
\cmidrule(lr){2-5} \cmidrule(lr){6-9}
Backbone & 1\,m & 3\,m & \textbf{5\,m} & 10\,m & 1\,m & 3\,m & \textbf{5\,m} & 10\,m \\
\midrule
Text2Loc & \SCARresult{19.10}{\textbf{(+2.54)}} & \SCARresult{37.67}{\textbf{(+6.36)}} & \SCARresult{54.68}{\textbf{(+10.05)}} & \SCARresult{62.43}{\textbf{(+11.63)}} & \SCARresult{7.76}{\textbf{(+1.72)}} & \SCARresult{24.14}{\textbf{(+6.03)}} & \SCARresult{26.72}{\textbf{(+6.90)}} & \SCARresult{30.17}{\textbf{(+6.90)}} \\
Text2Loc++ & \SCARresult{19.15}{\textbf{(+3.58)}} & \SCARresult{35.59}{\textbf{(+8.44)}} & \SCARresult{53.36}{\textbf{(+13.43)}} & \SCARresult{60.69}{\textbf{(+16.39)}} & \SCARresult{3.45}{(0.00)} & \SCARresult{14.66}{\textbf{(+6.03)}} & \SCARresult{16.38}{\textbf{(+6.90)}} & \SCARresult{19.83}{\textbf{(+8.62)}} \\
MNCL & \SCARresult{7.43}{\textbf{(+1.12)}} & \SCARresult{34.55}{\textbf{(+4.03)}} & \SCARresult{52.47}{\textbf{(+6.35)}} & \SCARresult{64.80}{\textbf{(+8.43)}} & \SCARresult{4.31}{\textbf{(+4.31)}} & \SCARresult{13.79}{\textbf{(+9.48)}} & \SCARresult{14.66}{\textbf{(+10.34)}} & \SCARresult{18.10}{\textbf{(+12.07)}} \\
MambaPlace & \SCARresult{7.24}{\textbf{(+0.49)}} & \SCARresult{31.68}{\textbf{(+4.38)}} & \SCARresult{47.71}{\textbf{(+7.04)}} & \SCARresult{60.21}{\textbf{(+9.66)}} & \SCARresult{4.31}{\textbf{(+0.86)}} & \SCARresult{15.52}{\textbf{(+2.59)}} & \SCARresult{19.83}{\textbf{(+2.59)}} & \SCARresult{24.14}{\textbf{(+3.45)}} \\
Des4Pos (reimpl.) & \SCARresult{6.27}{\textbf{(+0.26)}} & \SCARresult{23.79}{\textbf{(+3.24)}} & \SCARresult{41.44}{\textbf{(+5.42)}} & \SCARresult{59.76}{\textbf{(+8.28)}} & \SCARresult{0.00}{(0.00)} & \SCARresult{15.52}{\textbf{(+6.90)}} & \SCARresult{23.28}{\textbf{(+10.34)}} & \SCARresult{27.59}{\textbf{(+13.79)}} \\
\bottomrule
\end{tabular*}
\endgroup

\par\smallskip
\begin{minipage}{\textwidth}\footnotesize
Each comparison uses the same backbone, candidates, and Fine predictions. Differences are computed before rounding. Des4Pos denotes the corrected reimplementation. All TEST results are exploratory. $^{\dagger}$The MambaPlace$\times$PandaSet ep2 result is a protocol reference.
\end{minipage}
\end{table*}

\begin{table*}[!t]
\centering
\caption{TEST Cell R@1, R@3, R@5, and R@10 (\%) of SCAR. Each cell shows the recall above and the difference from Original (pp) below.}
\label{tab:cell}
\small
\setlength{\tabcolsep}{2pt}
\renewcommand{\arraystretch}{1.12}
\begingroup
\renewcommand{\SCARresult}[2]{\shortstack[c]{#1\\{\footnotesize #2}}}
\begin{tabular*}{\textwidth}{@{\extracolsep{\fill}}lcccccccccccc@{}}
\toprule
 & \multicolumn{4}{c}{\textbf{KITTI}} & \multicolumn{4}{c}{\textbf{S3DIS}} & \multicolumn{4}{c}{\textbf{PandaSet}} \\
\cmidrule(lr){2-5} \cmidrule(lr){6-9} \cmidrule(lr){10-13}
Backbone & R@1 & R@3 & R@5 & R@10 & R@1 & R@3 & R@5 & R@10 & R@1 & R@3 & R@5 & R@10 \\
\midrule
Text2Loc & \SCARresult{43.74}{\textbf{(+13.90)}} & \SCARresult{70.00}{\textbf{(+19.85)}} & \SCARresult{76.99}{\textbf{(+16.82)}} & \SCARresult{81.33}{\textbf{(+10.01)}} & \SCARresult{41.70}{\textbf{(+27.31)}} & \SCARresult{63.84}{\textbf{(+32.10)}} & \SCARresult{72.51}{\textbf{(+25.83)}} & \SCARresult{85.42}{\textbf{(+20.66)}} & \SCARresult{52.35}{\textbf{(+7.30)}} & \SCARresult{77.40}{\textbf{(+5.20)}} & \SCARresult{85.40}{\textbf{(+2.40)}} & \SCARresult{92.90}{\textbf{(+2.80)}} \\
Text2Loc++ & \SCARresult{45.02}{\textbf{(+10.02)}} & \SCARresult{72.36}{\textbf{(+14.74)}} & \SCARresult{80.15}{\textbf{(+13.26)}} & \SCARresult{85.02}{\textbf{(+8.09)}} & \SCARresult{48.15}{\textbf{(+26.20)}} & \SCARresult{70.11}{\textbf{(+24.54)}} & \SCARresult{82.29}{\textbf{(+24.91)}} & \SCARresult{89.11}{\textbf{(+18.08)}} & \SCARresult{49.10}{\textbf{(+7.95)}} & \SCARresult{69.15}{\textbf{(+4.45)}} & \SCARresult{83.20}{\textbf{(+9.00)}} & \SCARresult{91.75}{\textbf{(+2.65)}} \\
MNCL & \SCARresult{47.41}{\textbf{(+3.51)}} & \SCARresult{76.62}{\textbf{(+9.77)}} & \SCARresult{84.19}{\textbf{(+9.38)}} & \SCARresult{88.80}{\textbf{(+5.77)}} & \SCARresult{39.30}{\textbf{(+31.18)}} & \SCARresult{60.15}{\textbf{(+38.38)}} & \SCARresult{68.82}{\textbf{(+35.06)}} & \SCARresult{77.86}{\textbf{(+25.46)}} & \SCARresult{47.85}{\textbf{(+1.80)}} & \SCARresult{74.20}{\textbf{(+5.95)}} & \SCARresult{82.45}{\textbf{(+5.10)}} & \SCARresult{89.55}{\textbf{(+2.50)}} \\
MambaPlace$^{\dagger}$ & \SCARresult{40.83}{\textbf{(+16.28)}} & \SCARresult{65.34}{\textbf{(+21.56)}} & \SCARresult{72.03}{\textbf{(+18.83)}} & \SCARresult{76.26}{\textbf{(+11.79)}} & \SCARresult{42.25}{\textbf{(+28.78)}} & \SCARresult{64.76}{\textbf{(+33.03)}} & \SCARresult{73.80}{\textbf{(+31.55)}} & \SCARresult{82.66}{\textbf{(+23.99)}} & \SCARresult{29.50}{\textbf{(+9.70)}} & \SCARresult{46.90}{\textbf{(+11.95)}} & \SCARresult{57.35}{\textbf{(+11.55)}} & \SCARresult{73.20}{\textbf{(+14.85)}} \\
Des4Pos (reimpl.) & \SCARresult{43.11}{\textbf{(+19.44)}} & \SCARresult{65.72}{\textbf{(+23.06)}} & \SCARresult{71.67}{\textbf{(+20.11)}} & \SCARresult{75.35}{\textbf{(+12.07)}} & \SCARresult{39.11}{\textbf{(+23.06)}} & \SCARresult{61.44}{\textbf{(+27.31)}} & \SCARresult{72.51}{\textbf{(+29.15)}} & \SCARresult{82.10}{\textbf{(+18.82)}} & \SCARresult{41.55}{\textbf{(+13.45)}} & \SCARresult{61.70}{\textbf{(+8.10)}} & \SCARresult{72.05}{\textbf{(+7.85)}} & \SCARresult{83.25}{\textbf{(+6.35)}} \\
\bottomrule
\end{tabular*}
\par\medskip
\begin{tabular*}{\textwidth}{@{\extracolsep{\fill}}lcccccccc@{}}
\toprule
 & \multicolumn{4}{c}{\textbf{nuScenes}} & \multicolumn{4}{c}{\textbf{NPM3D}} \\
\cmidrule(lr){2-5} \cmidrule(lr){6-9}
Backbone & R@1 & R@3 & R@5 & R@10 & R@1 & R@3 & R@5 & R@10 \\
\midrule
Text2Loc & \SCARresult{62.19}{\textbf{(+11.58)}} & \SCARresult{76.31}{\textbf{(+9.78)}} & \SCARresult{81.16}{\textbf{(+7.32)}} & \SCARresult{87.42}{\textbf{(+4.39)}} & \SCARresult{31.03}{\textbf{(+6.90)}} & \SCARresult{53.45}{\textbf{(+9.48)}} & \SCARresult{76.72}{\textbf{(+15.52)}} & \SCARresult{95.69}{\textbf{(+16.38)}} \\
Text2Loc++ & \SCARresult{60.56}{\textbf{(+16.82)}} & \SCARresult{76.52}{\textbf{(+15.22)}} & \SCARresult{80.29}{\textbf{(+11.09)}} & \SCARresult{85.57}{\textbf{(+6.62)}} & \SCARresult{19.83}{\textbf{(+8.62)}} & \SCARresult{43.10}{\textbf{(+6.03)}} & \SCARresult{71.55}{\textbf{(+20.69)}} & \SCARresult{93.10}{\textbf{(+14.66)}} \\
MNCL & \SCARresult{65.28}{\textbf{(+8.97)}} & \SCARresult{81.26}{\textbf{(+8.59)}} & \SCARresult{86.82}{\textbf{(+6.27)}} & \SCARresult{91.61}{\textbf{(+4.11)}} & \SCARresult{19.83}{\textbf{(+13.79)}} & \SCARresult{45.69}{\textbf{(+22.41)}} & \SCARresult{62.93}{\textbf{(+25.86)}} & \SCARresult{95.69}{\textbf{(+39.66)}} \\
MambaPlace & \SCARresult{60.11}{\textbf{(+9.82)}} & \SCARresult{76.63}{\textbf{(+9.00)}} & \SCARresult{81.40}{\textbf{(+6.75)}} & \SCARresult{87.71}{\textbf{(+4.12)}} & \SCARresult{24.14}{\textbf{(+3.45)}} & \SCARresult{45.69}{\textbf{(+6.03)}} & \SCARresult{66.38}{\textbf{(+9.48)}} & \SCARresult{93.97}{\textbf{(+18.10)}} \\
Des4Pos (reimpl.) & \SCARresult{60.11}{\textbf{(+8.48)}} & \SCARresult{77.98}{\textbf{(+7.61)}} & \SCARresult{84.21}{\textbf{(+7.31)}} & \SCARresult{89.84}{\textbf{(+3.62)}} & \SCARresult{27.59}{\textbf{(+13.79)}} & \SCARresult{46.55}{\textbf{(+15.52)}} & \SCARresult{62.93}{\textbf{(+18.10)}} & \SCARresult{86.21}{\textbf{(+23.28)}} \\
\bottomrule
\end{tabular*}
\endgroup

\par\smallskip
\begin{minipage}{\textwidth}\footnotesize
Differences are computed before rounding within the same backbone and dataset. Des4Pos denotes the corrected reimplementation. All TEST results are exploratory. $^{\dagger}$The MambaPlace$\times$PandaSet ep2 result is a protocol reference.
\end{minipage}
\end{table*}

\begin{table*}[!t]
\centering
\caption{Same-pool TEST comparison of Semantic Matching, Rank Fusion, FSR, and SCAR. Each entry reports Top-1@5\,m / Cell R@1 (\%).}
\label{tab:fullcontrols}
\small
\setlength{\tabcolsep}{3pt}
\renewcommand{\arraystretch}{1.07}
\begin{tabular*}{\textwidth}{@{\extracolsep{\fill}}llccccc@{}}
\toprule
Dataset & Backbone & Original & Semantic Matching & Rank Fusion & FSR & SCAR \\
\midrule
KITTI & Text2Loc & 35.14 / 29.84 & 49.05 / 41.28 & 54.58 / 44.62 & 58.65 / \textbf{46.51} & \textbf{59.45} / 43.74 \\
 & Text2Loc++ & 43.72 / 35.00 & 56.45 / 44.98 & 59.05 / 46.05 & 63.36 / \textbf{47.51} & \textbf{64.15} / 45.02 \\
 & MNCL & 49.59 / 43.90 & 54.90 / 46.01 & 57.56 / 46.62 & 62.00 / \textbf{48.19} & \textbf{63.82} / 47.41 \\
 & MambaPlace & 31.98 / 24.55 & 45.31 / 35.74 & 53.72 / 42.36 & \textbf{57.65} / \textbf{44.18} & 57.07 / 40.83 \\
 & Des4Pos (reimpl.) & 29.37 / 23.67 & 46.03 / 36.71 & 53.01 / 42.67 & 57.41 / \textbf{44.37} & \textbf{57.84} / 43.11 \\
\addlinespace[2pt]
S3DIS & Text2Loc & 13.84 / 14.39 & 28.23 / 29.89 & 36.90 / 38.38 & \textbf{44.83} / \textbf{46.49} & 41.33 / 41.70 \\
 & Text2Loc++ & 20.66 / 21.96 & 32.10 / 34.32 & 38.56 / 40.41 & 45.20 / 47.42 & \textbf{45.94} / \textbf{48.15} \\
 & MNCL & 7.93 / 8.12 & 28.97 / 30.26 & 34.69 / 35.79 & 37.64 / 38.93 & \textbf{38.38} / \textbf{39.30} \\
 & MambaPlace & 11.99 / 13.47 & 29.52 / 31.92 & 35.42 / 37.08 & 37.45 / 39.30 & \textbf{40.77} / \textbf{42.25} \\
 & Des4Pos (reimpl.) & 14.58 / 16.05 & 30.44 / 33.03 & 34.32 / 36.16 & \textbf{38.38} / \textbf{40.41} & 37.82 / 39.11 \\
\addlinespace[2pt]
PandaSet & Text2Loc & 40.95 / 45.05 & \textbf{48.75} / \textbf{52.95} & 43.75 / 46.75 & 40.50 / 43.10 & 44.95 / 52.35 \\
 & Text2Loc++ & 37.35 / 41.15 & 42.30 / 47.30 & 41.25 / 44.85 & 37.90 / 41.35 & \textbf{44.30} / \textbf{49.10} \\
 & MNCL & 42.20 / 46.05 & 30.25 / 34.10 & 35.10 / 38.75 & 35.40 / 38.95 & \textbf{42.05} / \textbf{47.85} \\
 & MambaPlace$^{\dagger}$ & 13.90 / 19.80 & 19.85 / 28.85 & 22.30 / 30.90 & \textbf{24.15} / \textbf{31.75} & 22.15 / 29.50 \\
 & Des4Pos (reimpl.) & 20.35 / 28.10 & 29.50 / 40.00 & \textbf{32.15} / \textbf{41.80} & 30.45 / 37.30 & 31.55 / 41.55 \\
\addlinespace[2pt]
nuScenes & Text2Loc & 44.63 / 50.61 & 50.93 / 58.08 & 47.26 / 54.22 & 42.58 / 49.01 & \textbf{54.68} / \textbf{62.19} \\
 & Text2Loc++ & 39.93 / 43.74 & 51.20 / 58.72 & 47.21 / 54.60 & 44.32 / 51.30 & \textbf{53.36} / \textbf{60.56} \\
 & MNCL & 46.12 / 56.31 & 41.36 / 52.74 & 35.54 / 45.90 & 32.18 / 41.66 & \textbf{52.47} / \textbf{65.28} \\
 & MambaPlace & 40.67 / 50.29 & 41.15 / 53.24 & 36.44 / 47.08 & 33.32 / 42.46 & \textbf{47.71} / \textbf{60.11} \\
 & Des4Pos (reimpl.) & 36.03 / 51.63 & 37.77 / 55.40 & 34.34 / 50.07 & 31.03 / 45.40 & \textbf{41.44} / \textbf{60.11} \\
\addlinespace[2pt]
NPM3D & Text2Loc & 19.83 / 24.14 & \textbf{27.59} / \textbf{33.62} & \textbf{27.59} / \textbf{33.62} & \textbf{27.59} / \textbf{33.62} & 26.72 / 31.03 \\
 & Text2Loc++ & 9.48 / 11.21 & \textbf{17.24} / \textbf{19.83} & \textbf{17.24} / \textbf{19.83} & \textbf{17.24} / \textbf{19.83} & 16.38 / \textbf{19.83} \\
 & MNCL & 4.31 / 6.03 & 9.48 / 13.79 & 9.48 / 13.79 & 9.48 / 13.79 & \textbf{14.66} / \textbf{19.83} \\
 & MambaPlace & 17.24 / 20.69 & \textbf{19.83} / \textbf{24.14} & \textbf{19.83} / \textbf{24.14} & \textbf{19.83} / \textbf{24.14} & \textbf{19.83} / \textbf{24.14} \\
 & Des4Pos (reimpl.) & 12.93 / 13.79 & \textbf{23.28} / \textbf{27.59} & \textbf{23.28} / \textbf{27.59} & \textbf{23.28} / \textbf{27.59} & \textbf{23.28} / \textbf{27.59} \\
\bottomrule
\end{tabular*}

\par\smallskip
\begin{minipage}{\textwidth}\footnotesize
Each configuration shares its candidate pool and Fine predictions across the four rankings. Supplement Tables S4 and S5 give the additional distance thresholds and cell ranks. Des4Pos denotes the corrected reimplementation. All TEST results are exploratory. $^{\dagger}$The MambaPlace$\times$PandaSet ep2 result is a protocol reference.
\end{minipage}
\end{table*}

After setting MambaPlace$\times$PandaSet aside as a protocol reference, exploratory TEST point estimates on the remaining 24 configurations show that SCAR's Top-1@5\,m is above Original in 23 and Cell R@1 is above Original in all 24. The only configuration below Original on Top-1@5\,m is MNCL$\times$PandaSet, by $-0.15$\,pp. For example, Text2Loc on KITTI rises from 35.14\% to 59.45\%, and on S3DIS from 13.84\% to 41.33\%.

Relative to same-backbone FSR, SCAR's localization gain varies by configuration. For corrected Des4Pos on PandaSet, the Top-1@5\,m difference is $+1.10$\,pp with a 95\% paired scene-cluster interval of $[-1.35,3.40]$. On nuScenesFull, the difference is $+10.41$\,pp with interval $[7.39,13.62]$. Table~\ref{tab:fullcontrols} gives the primary metric values, and Supplement Table S6 summarizes the configuration-level comparisons.

\subsection{Complementarity Between Learned Fusion and Strong Controls}
On MambaPlace$\times$nuScenes, Top-1@5\,m for Original, Semantic Matching, FSR, and SCAR is 40.67\%, 41.15\%, 33.32\%, and 47.71\%, respectively; SCAR improves over Original by 7.04\,pp and over the stronger Semantic Matching control by 6.56\,pp. This configuration reuses its existing frozen weights and cache.

An additional scene-cluster sensitivity analysis uses 210 scenes and retains all queries within each scene. SCAR's Top-1@5\,m intervals relative to Original, Semantic Matching, and FSR are $[5.24,8.92]$, $[4.93,8.36]$, and $[11.63,17.41]$\,pp, respectively, all positive; the original pose-cluster intervals are retained as well.

On MambaPlace$\times$S3DIS, SCAR is 40.77\%, Original is 11.99\%, and FSR is 37.45\%; SCAR improves over Original by 28.78\,pp and over FSR by 3.32\,pp. The room-cluster interval is $[0.18,6.83]$; the 2.95\,pp Cell R@1 difference has interval $[-0.18,6.48]$. The positive localization difference in this configuration is not accompanied by a cell-retrieval advantage.

Intervals are interpreted in the query or clustering unit recorded by their frozen protocol and are not substituted across units. The rule controls provide a reference against which additional gain can be measured. On Text2Loc$\times$PandaSet, SCAR improves over Original by 4.00\,pp, while Semantic Matching is 3.80\,pp above SCAR on Top-1@5\,m.

On Text2Loc$\times$S3DIS, SCAR improves Top-1@5\,m over Original by 27.49\,pp. Relative to FSR, SCAR is 3.51 and 4.80\,pp lower in Top-1@5\,m and Cell R@1, with paired room-cluster intervals of $[-6.99,-0.18]$ and $[-8.86,-0.92]$. On MNCL$\times$PandaSet, Original and SCAR achieve 42.20\% and 42.05\% Top-1@5\,m, both above the three rule-based rerankings.

\subsection{Localization Behavior and Ranking Depth on KITTI}
On KITTI, SCAR improves Top-1@5\,m over Original by 24.31, 20.43, and 14.23\,pp for Text2Loc, Text2Loc++, and MNCL, respectively, and Cell R@1 by 13.90, 10.02, and 3.51\,pp. Relative to FSR, the three localization differences are 0.80, 0.79, and 1.82\,pp, with pose-cluster intervals $[0.17,1.44]$, $[0.16,1.44]$, and $[1.09,2.51]$ based on 3,932 pose clusters from three scenes; Cell R@1 is 2.77, 2.49, and 0.77\,pp lower, and the first two intervals exclude zero.

On MambaPlace$\times$KITTI, relative to FSR, SCAR improves Top-10@5\,m, Cell R@5, and R@10 by 0.48, 0.79, and 0.75\,pp. Under the same comparison, Top-1@5\,m and Cell R@1 are 0.58 and 3.35\,pp lower, with paired pose-cluster intervals $[-1.15,-0.03]$ and $[-3.96,-2.74]$. The direction of change in this configuration depends on ranking depth.

\begin{figure*}[!t]
\centering
\includegraphics[width=0.95\textwidth]{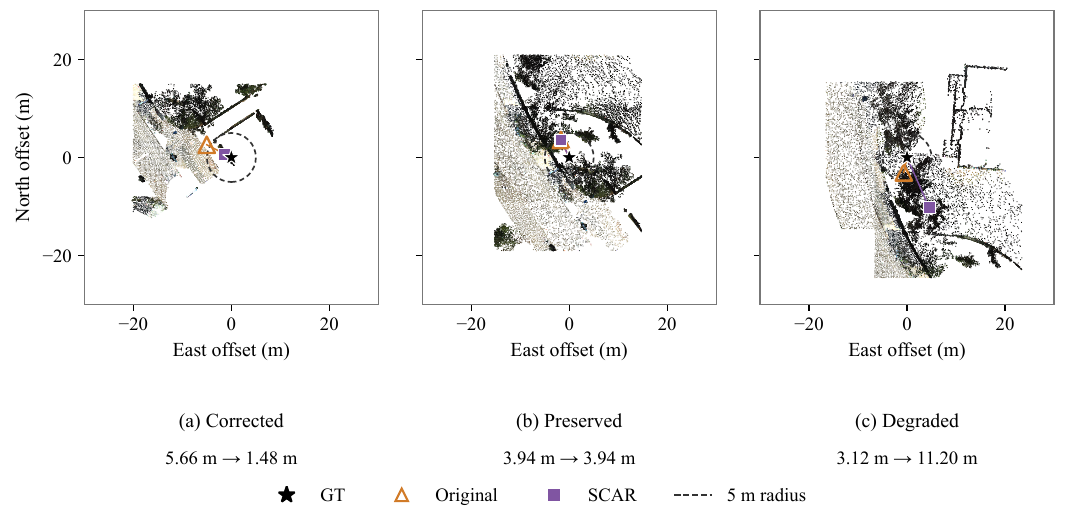}
\caption{Real corrected Des4Pos$\times$KITTI VAL cases, shown from left to right: (a) correction, (b) preservation, and (c) degradation. Stars denote GT, orange triangles Original predictions, and purple squares SCAR predictions; dashed circles have a 5\,m radius. Coordinates are relative to GT, with the same scale in all panels. Errors below each panel are Original$\rightarrow$SCAR. The two predictions coincide in (b). Case identities, query descriptions, and selection details are given in Supplement S4.3.}
\label{fig:cases}
\end{figure*}

The two metric families follow from the hypothesis-selection problem: a lower-error prediction can be selected from a cell other than the annotated GT cell. Fig.~\ref{fig:cases} illustrates first-rank selection on three fixed corrected Des4Pos$\times$KITTI VAL cases from scene 0010, using the formal seed-0 epoch-26 head. Query 0002 changes from an Original error of 5.66\,m to 1.48\,m after reranking; query 0028 retains the same cell and Fine point at 3.94\,m; query 0071 changes from 3.12\,m to 11.20\,m. The panels retain the original object-point RGB and prediction coordinates. They are the first eligible cases within predefined correction, preservation, and degradation strata in the locally available cell subset. GT centers the display and defines error only; it is not used by the inference features. The original query descriptions and candidate identities are recorded in Supplement S4.3. These examples illustrate selection outcomes; controlled support-pathway comparisons follow in the next section.

\section{Spatial Evidence and the Hypothesis-Selection Mechanism}\label{sec:mechanism}
\subsection{Does the Readout Position Affect Selection?}
\begin{figure*}[!t]
\centering
\includegraphics[width=0.95\textwidth]{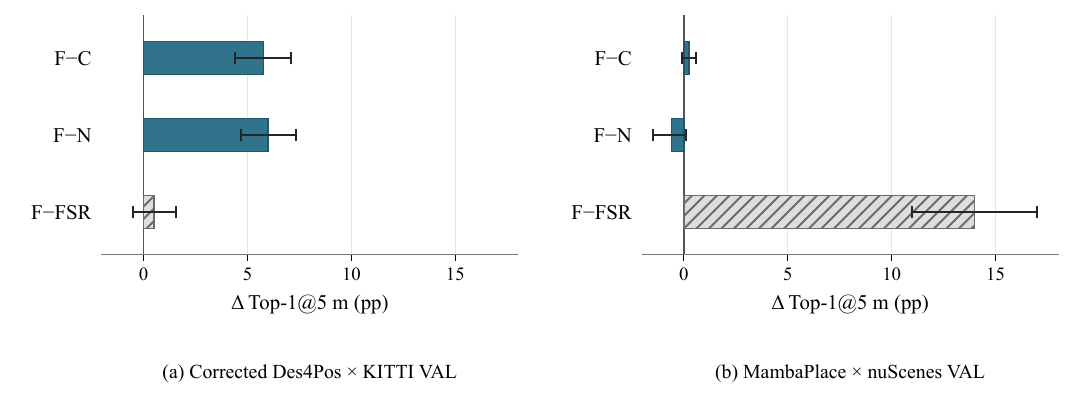}
\caption{VAL Top-1@5\,m differences between F and each control: (a) corrected Des4Pos$\times$KITTI and (b) MambaPlace$\times$nuScenesFull. Bars show point estimates; whiskers are paired 95\% cluster-bootstrap intervals. Filled bars compare trained F/C/N arms, while hatched bars compare the complete F scoring scheme with training-free FSR. Both panels use scorer seed 0 and the same horizontal scale. Complete results are in Supplement Tables S8 and S9.}
\label{fig:ablation}
\end{figure*}

To test the role of Fine-position readout, F/C/N controls are run on two frozen input sets. F is the complete 23-D feature with support read at the Fine prediction point; C replaces the support readings of columns 12--16 (zero-based) with center readings; N masks columns 12--16 after normalization, removing Fine support and the derived ranking pathway while retaining center support (columns 7/8), center rank, candidate pool, and other shared evidence. The three arms share candidates, Fine geometry, labels, normalization pipeline, network architecture, and epoch-selection rule, and differ only in the support pathway. Fig.~\ref{fig:ablation} summarizes Top-1@5\,m differences (F$-$C, F$-$N, and F$-$FSR) for the two configurations. Corrected Des4Pos$\times$KITTI uses 3,187 VAL queries in 1,116 pose clusters; MambaPlace$\times$nuScenesFull uses 7,635 VAL queries in 209 scene clusters. Seed 0 is primary in both configurations. The 95\% intervals use 2,000 paired cluster-bootstrap resamples (seed 123), with the registered cluster unit for each configuration. Complete readings are in Supplement Tables S8 and S9.

In Fig.~\ref{fig:ablation}(a), Fine readout improves over center readout by 5.77\,pp (95\% interval $[4.42,7.09]$) and over removing the whole pathway by 5.99\,pp ($[4.68,7.32]$), with consistent direction across three seeds. The corresponding Cell R@1 differences are $-1.60$ and $-0.38$\,pp, both with intervals crossing zero. The C$-$N localization difference is $+0.22$\,pp ($[-0.69,1.14]$). Relative to FSR, F's localization difference is $+0.50$\,pp ($[-0.53,1.57]$), and its Cell R@1 difference is $-3.83$\,pp ($[-5.31,-2.38]$).

\subsection{Complete Scoring Scheme Versus Rule-Based Ranking}
The same F/C/N protocol is applied to the frozen TRAIN/VAL features of MambaPlace$\times$nuScenes (7,635 VAL queries, 209 scene clusters). The network, loss, optimizer, and epoch-selection rule are unchanged; Supplement Table S9 gives complete seed-0 results with scene-cluster paired bootstrap intervals, and Fig.~\ref{fig:ablation}(b) gives Top-1@5\,m differences.

In this configuration, the complete scoring scheme improves Top-1@5\,m over FSR by $+13.98$\,pp (95\% interval $[10.97,17.01]$) and Cell R@1 by $+17.22$\,pp ($[13.72,20.71]$), with consistent direction across all nine trained heads. Among the support-pathway controls, F$-$C is $+0.25$\,pp ($[-0.10,0.60]$) and F$-$N is $-0.58$\,pp ($[-1.48,0.12]$), with the direction varying by seed. C$-$N is $-0.83$\,pp ($[-1.72,-0.11]$) for seed 0, and the direction reverses for the other two seeds. The two VAL studies use different backbones, query distributions, and feature constructions. The $+13.98$\,pp figure compares the complete scoring scheme with FSR and therefore includes feature construction and supervision; it is not an isolated estimate of the contribution of learned ranking.

\subsection{Effect of the Supervision Label on Position Selection}
The supervision target determines what the scorer learns. To isolate the label definition, we fix the backbone, Top-25 candidates, Fine coordinates, directional gate $m\geq4$, and scorer architecture on corrected Des4Pos$\times$KITTI and change only each candidate's positive label. Both arms use the same 24,908 TRAIN queries, 23-D features, and normalization pipeline, normalized by the number of positives per query; 8,396 queries have identical positive sets under the two labels. The localization-success label is positive when the candidate cell belongs to the same scene as the ground-truth pose and the candidate's Fine-predicted position is within 5\,m of that pose. the candidate to be in the same scene as the actual Fine position and within 5\,m; the GT-cell label requires the candidate ID to equal the target cell ID.

For seed 0, the localization-success label gives 60.81\% VAL Top-1@5\,m and the GT-cell label 58.64\%, a difference of $+2.17$\,pp (95\% CI $[0.85,3.44]$); Cell R@1 for the same scorer is 44.74\% and 53.34\%, a difference of $-8.60$\,pp ($[-10.34,-6.90]$). For seeds 1 and 2, Top-1@5\,m differences are $+1.51$ and $+1.57$\,pp, and Cell R@1 differences are $-6.46$ and $-8.06$\,pp; the direction is consistent across all three seeds.

The two labels optimize different objects: the localization-success label directly rewards a first-ranked coordinate within 5\,m, whereas the GT-cell label requires the first rank to fall in the correct map cell. The former improves coordinate precision and the latter improves cell-level retrieval; in this configuration they exhibit a clear trade-off rather than one label improving every metric. This is a VAL development-set ablation and does not change the main protocol's localization-success supervision; it compares supervision targets within the same learned scorer and is not used to separate learned ranking from FSR. Complete three-seed readings are in Supplement Table S10.

\subsection{Directional Gate and Candidate-Evidence Distribution}
The directional gate determines how much effective information is available in different domains. Each query in ParisLillePose/NPM3D has only three original descriptions, so the number of effective relations cannot reach four and the directional channel is identically zero under the frozen rule. In these configurations, Semantic Matching, Rank Fusion, and FSR have identical TEST readings; the learned scorer can produce ranking differences only from the remaining available evidence.

\begin{figure*}[!t]
\centering
\includegraphics[width=0.95\textwidth]{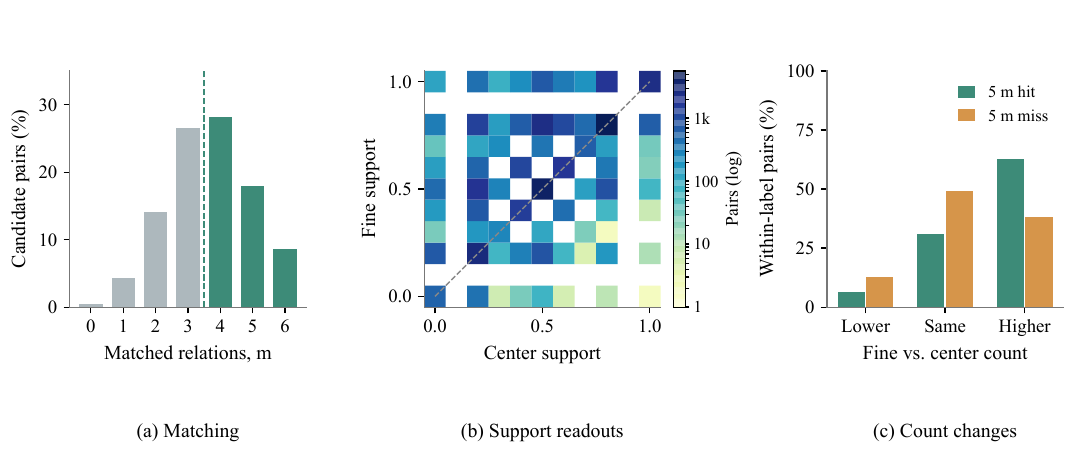}
\caption{Spatial-support distributions on MNCL$\times$KITTI VAL. (a) Effective matched-relation count $m$, with the dashed line marking the $m\geq4$ gate. (b) Joint histogram of center and Fine support among active-gate pairs; color indicates pair count on a logarithmic scale, white bins are empty, and the diagonal marks equal support. (c) Changes in satisfied-relation count at the Fine point relative to the center, normalized separately within 5\,m-hit and 5\,m-miss candidate pairs. These are descriptive distributions, not ranking ablations.}
\label{fig:gate}
\end{figure*}

Fig.~\ref{fig:gate} describes the two readouts on 79,675 MNCL$\times$KITTI VAL candidate pairs (3,187 queries $\times$ 25 candidates). The $m\geq4$ gate is active for 43,476 pairs (54.57\%). Effective relation count and the presence-based matching count disagree on 9,079 pairs, so these counts are kept distinct. Among active-gate pairs, the Fine satisfied count is higher than, equal to, or lower than the center count for 17,986, 20,369, and 5,121 pairs, respectively. Panel (b) uses the original 0.1-wide bins for the two $C/m$ readouts; panel (c) normalizes the count changes separately over 5,671 localization-hit pairs and 37,805 miss pairs. Here a hit denotes a same-scene Fine prediction within 5\,m of GT. The figure describes correlated candidate pairs sharing queries and cells; the controlled F/C/N contrasts, rather than this distribution, quantify changes in ranking performance. These MNCL statistics and the following corrected Des4Pos gate ablation refer to different backbone-specific candidate sets.

On corrected Des4Pos$\times$KITTI VAL, we fix the localization label and relax the gate from $m\geq4$ to $m\geq1$, $m\geq2$, and $m\geq3$; the paired intervals of Top-1@5\,m across three seeds all cross zero (point differences from $-0.16$ to $+0.53$\,pp), and the Cell R@1 interval for $m\geq2$ is above zero for two seeds and crosses zero for the third. In this VAL candidate set, $m<4$ accounts for 44,537/79,675 (55.90\%). No stable 5\,m localization gain is observed; complete results are in Supplement Table S11.

\section{Efficiency Analysis}\label{sec:efficiency}
\subsection{Inference Cost and Candidate Budget}
The existing formal latency measurements were run serially on the same RTX 4080 SUPER with 128 fixed KITTI VAL queries, batch 1, FP32, 20 warm-up queries, and five repetitions, under eval mode and no\_grad, synchronizing the GPU at stage boundaries. The table aggregates 128 queries and 640 observations.

The main table uses a warm-cache protocol covering gallery features, pre-organized candidate objects, and per-cell auxiliary information; offline index construction and cold start are reported separately. The native pipeline, the fixed Top-25 base pipeline, FSR, and SCAR are reported separately. The existing scripts record the sum of stage times for text encoding, gallery retrieval, Fine, spatial evidence, feature construction, and the scoring network, but do not yet cover final sorting and coordinate return; Table~\ref{tab:latency} is therefore a stage-wise computational latency, not end-to-end wall time.

\begin{table}[!t]
\centering
\caption{KITTI fixed-VAL-query median stage-wise computational latency (ms) under warm cache. Text2Loc native K10/fixed K25 is 126.40/275.99; MambaPlace was not measured. The Des4Pos row is the historical implementation; its corrected reimplementation was not measured.}
\label{tab:latency}
\footnotesize
\setlength{\tabcolsep}{4pt}
\begin{tabular}{@{}lcccc@{}}
\toprule
Method & Native K10 & Same-pool K25 & FSR & SCAR \\
\midrule
Text2Loc & 126.40 & 275.99 & 279.68 & 276.39 \\
Text2Loc++ & 93.69 & 207.64 & 212.77 & 214.89 \\
MNCL & 252.87 & 584.45 & 588.88 & 592.78 \\
MambaPlace & \multicolumn{4}{c}{Not measured} \\
Des4Pos (historical impl.) & 100.49 & 191.98 & 199.65 & 196.25 \\
\bottomrule
\end{tabular}
\end{table}

For the measured configurations in Table~\ref{tab:latency}, spatial-evidence construction, feature computation, and scoring together take 4.73--5.70\,ms, of which the scoring network takes 0.36--0.52\,ms. The range includes the historical Des4Pos implementation. The Fine-forward cost of extending the native 10 candidates to 25 is reported separately as a budget change.

The 23-D scoring network contains 3,649 parameters. Counting 1 MAC as 2 FLOPs, the three linear maps process 25 candidates with 88,800 MACs, i.e., 0.0001776 GFLOPs per query. This value covers only the scorer's linear maps.

\subsection{Training and Deployment Cost}
Efficiency also depends on the division between training and deployment. SCAR trains a small scorer on frozen backbone and candidate predictions and reuses gallery features and object auxiliary caches at deployment; backbone training, candidate export, offline index construction, and online querying should be measured separately. Table~\ref{tab:train} distinguishes wall time from summed per-epoch time by source and records missing logs as not recorded; training cost is not inferred from parameter counts.

\begin{table}[!t]
\centering
\caption{Stage training durations in the existing logs. Durations correspond to the original run environment and recipe of each configuration and exclude failed attempts and overall project development overhead.}
\label{tab:train}
\scriptsize
\setlength{\tabcolsep}{3pt}
\begin{tabular}{@{}lccp{2.45cm}@{}}
\toprule
Configuration & Coarse & Fine & Evidence type \\
\midrule
MambaPlace$\times$KITTI & 8 h 37 min 09 s & 14 h 03 min 26 s & Start/end wall clock \\
MambaPlace$\times$PandaSet & 1 h 08 min 26 s & 1 h 45 min 50 s & Session transcript; log not archived \\
MambaPlace$\times$nuScenes & 5 h 30 min 01 s & 8 h 18 min 01 s & Start/end wall clock \\
MambaPlace$\times$S3DIS & 27 min 22 s & 31 min 11 s & Start/end wall clock \\
MambaPlace$\times$NPM3D & 2 min 06 s & 3 min 52 s & Start/end wall clock \\
Text2Loc++$\times$S3DIS & 47 min 44 s & 45 min 23 s & Start/end wall clock \\
Text2Loc++$\times$NPM3D & 2 min 00 s & 3 min 51 s & Start/end wall clock \\
Text2Loc++$\times$nuScenes & Not recorded & 8 h 20 min 23 s & Fine start/end wall clock \\
Text2Loc$\times$NPM3D & 97 s & 197 s & Start/end wall clock \\
\bottomrule
\end{tabular}
\end{table}

\section{Discussion and Limitations}\label{sec:discussion}
SCAR uses object maps with category and color, template-based directional relations, and a fixed candidate pool. The scorer for each target domain is fitted on that domain's TRAIN split, and the results describe hypothesis selection under target-domain adaptation.

The fixed-hypothesis interface allows ranking changes to be compared within the same coordinate set, but it also bounds the achievable gain: regions outside the candidate pool cannot be recovered, and queries for which all candidates fail cannot be corrected by reranking alone. Enlarging the pool increases coverage while raising Fine-forward cost; this paper studies this trade-off with a fixed Top-25 budget.

KITTI TEST covers three scenes, and NPM3D TEST has 116 queries. Paired intervals are interpreted according to the resampling unit registered for each result. In NPM3D configurations where the directional support gate is not triggered, Semantic Matching, Rank Fusion, and FSR yield identical test readings.

Rule-based rankings provide a reference for measuring the increment of the learned scheme. The two VAL studies separately reveal the role of readout position and the performance of the complete scoring scheme relative to FSR. The latter comparison should not be read as the isolated contribution of learned ranking.

Several limitations remain. The label and gate ablations are VAL studies on a single backbone, and the main results are exploratory TEST readings rather than confirmatory evaluations. The scorer uses one primary initialization seed for the main table. Future work should repeat the readout and label ablations across backbones and independent test sets, and should measure deployment cost including all system components rather than stage-wise scorer overhead alone.

\section{Conclusion}
SCAR formulates the final output of coarse-to-fine localization as a selection problem over a fixed candidate pool. The scorer reads directional support at the candidate's existing Fine prediction point and combines it with center, retrieval, and geometric evidence to reorder localization hypotheses. After setting MambaPlace$\times$PandaSet aside as a protocol reference, exploratory TEST point estimates on the remaining 24 configurations show that SCAR's Top-1@5\,m is above Original in 23 and Cell R@1 is above Original in all 24. VAL controls show that the Fine readout improves localization success on corrected Des4Pos$\times$KITTI, and that the complete scoring scheme outperforms FSR on MambaPlace$\times$nuScenesFull. These results establish candidate selection as a learnable and independently evaluable stage of coarse-to-fine localization.

\section*{Supplementary Material}
The supplementary document provides the 23-D feature specification (S1), complete TEST readings (S2), provenance and reading rules (S3), reproducibility and case provenance (S4), and VAL diagnostics and ablations (S5), including Tables S1--S11 cited above.

\bibliographystyle{IEEEtran}
\bibliography{Bibliography}

\end{document}


\maketitle

\noindent This supplement accompanies the manuscript \emph{SCAR: Bridging Coarse Retrieval and Fine Localization through Candidate Reranking}. Its content is translated from the verified Chinese v13 source and uses only the frozen experimental records reported there. No experiment was rerun, no TEST split was read again, and no number was transferred from another manuscript version. Section~S1 specifies the 23-D representation. Section~S2 gives the complete exploratory TEST readings. Section~S3 records provenance and reading rules. Section~S4 documents reproducibility, the Des4Pos reimplementation, case provenance, and the identity of the efficiency row. Section~S5 reports the VAL diagnostics and controlled ablations. Section~S6 reports Original localization rates.

\section*{S1\quad The 23-D Candidate Representation}

Each candidate is described by 23 features arranged in seven evidence groups. The groups follow the order in which evidence is judged: coarse retrieval and semantic presence; candidate organization and center support; the Fine prediction's displacement from the center; spatial support read at the Fine point; rank fusion; object and neighborhood geometry; and position relative to the list anchor. Table~\ref{tabS:features} lists the canonical column names, zero-based indices, definitions, and pre-standardization normalization.

The representation is shared by all configurations. The scorer is a small network mapping 23 dimensions to 64, then 32, then one score, with 3,649 parameters. All candidates of a query are scored by the same network, and the list is ordered by the resulting scores. Only the existing Fine coordinates are reranked; no new coordinate is predicted at inference.

\begin{table}[H]
\centering
\caption{The 23-D candidate representation. Indices are zero-based. The seven evidence groups contain 4, 5, 3, 3, 2, 4, and 2 dimensions. ``Cell units'' are the map-grid units used by the backbone; ``anchor'' is the first candidate in canonical P2 order.}
\label{tabS:features}
\footnotesize
\setlength{\tabcolsep}{4pt}
\begin{tabular}{@{}p{0.8cm}p{3.25cm}p{9.25cm}p{2.55cm}@{}}
\toprule
Idx & Column name & Definition & Normalization \\
\midrule
\multicolumn{4}{@{}l}{\textit{Group 1: coarse retrieval and semantic presence (4-D)}} \\
0 & \texttt{cos} & Cosine similarity between the query embedding and the candidate's coarse scene-text embedding. & raw \\
1 & \texttt{cos\_rank} & Rank of \texttt{cos} within the fixed candidate list. & $/(K-1)$ \\
2 & \texttt{sp} & Matched-presence fraction from the relational presence scorer. & $[0,1]$ \\
3 & \texttt{p0\_rank} & Rank in the presence-augmented P0 order defined by $r_P=\texttt{cos}+0.5\,\texttt{sp}$. & $/(K-1)$ \\
\midrule
\multicolumn{4}{@{}l@{}}{} \\
\multicolumn{4}{@{}l}{\textit{Group 2: candidate organization and center evidence (5-D)}} \\
4 & \texttt{p2\_rank} & Position after the directional gate and canonical reordering (the P2 order). Ties occur only inside the maximum-presence block. & $/(K-1)$ \\
5 & \texttt{is\_tied} & Indicator that the candidate belongs to the maximum-presence tie block. & $\{0,1\}$ \\
6 & \texttt{tied\_size} & Number of candidates in that tie block. & $/K$ \\
7 & \texttt{sc\_center} & Directional-relation support score evaluated at the cell center. & $[0,1]$ \\
8 & \texttt{sc\_center\_rank} & Rank of \texttt{sc\_center} within the candidate list. & $/(K-1)$ \\
\midrule
\multicolumn{4}{@{}l@{}}{} \\
\multicolumn{4}{@{}l}{\textit{Group 3: Fine displacement relative to the center (3-D)}} \\
9 & \texttt{off\_dx} & East displacement from the cell center to the Fine prediction. & cell units \\
10 & \texttt{off\_dy} & North displacement from the cell center to the Fine prediction. & cell units \\
11 & \texttt{off\_norm} & Euclidean length of $(\texttt{off\_dx},\texttt{off\_dy})$. & cell units \\
\midrule
\multicolumn{4}{@{}l@{}}{} \\
\multicolumn{4}{@{}l}{\textit{Group 4: Fine spatial support (3-D)}} \\
12 & \texttt{sc\_fine} & Directional-relation support score evaluated at the Fine prediction point. & $[0,1]$ \\
13 & \texttt{sat\_fine} & Proportion of query relations satisfied at the Fine point. & $[0,1]$ \\
14 & \texttt{sc\_fine\_rank} & Rank of \texttt{sc\_fine} within the candidate list. & $/(K-1)$ \\
\midrule
\multicolumn{4}{@{}l@{}}{} \\
\multicolumn{4}{@{}l}{\textit{Group 5: rank fusion (2-D)}} \\
15 & \texttt{borda} & Normalized Borda aggregate of \texttt{p2\_rank} and \texttt{sc\_fine\_rank}. & $/(2(K-1))$ \\
16 & \texttt{c\_rank} & Position in the canonical Borda order; this order defines candidate identity and cache order. & $/(K-1)$ \\
\midrule
\multicolumn{4}{@{}l@{}}{} \\
\multicolumn{4}{@{}l}{\textit{Group 6: object and neighborhood geometry (4-D)}} \\
17 & \texttt{n\_obj} & Number of matched objects retained in the candidate region. & $/40$ \\
18 & \texttt{n\_matched} & Number of query descriptions with at least one matched object. & $/n_q$ \\
19 & \texttt{dist\_nearest\_matched} & Distance from the Fine prediction to the nearest matched object. & cell units, capped at 3.0 \\
20 & \texttt{dist\_centroid\_matched} & Distance from the Fine prediction to the centroid of matched objects. & cell units, capped at 3.0 \\
\midrule
\multicolumn{4}{@{}l@{}}{} \\
\multicolumn{4}{@{}l}{\textit{Group 7: relative anchor position (2-D)}} \\
21 & \texttt{pred\_dist\_anchor} & Distance from the Fine prediction to the P2-top-1 predicted position. & cell units, capped at 3.0 \\
22 & \texttt{center\_dist\_anchor} & Distance from the cell center to the P2-top-1 cell center. & cell units, capped at 3.0 \\
\bottomrule
\end{tabular}
\begin{flushleft}
\footnotesize
Note: $K=25$ is the fixed candidate budget and $n_q$ is the number of descriptions for the query. Rank columns are divided by $K-1$; the tie-block size by $K$; the Borda aggregate by $2(K-1)$; object count by 40; matched-object count by $n_q$; and distance columns are capped at 3.0. After these constructions, every column is standardized with the TRAIN mean and standard deviation, with the standard deviation floored at $10^{-6}$. In the two configurations that received code-level auditing (corrected Des4Pos$\times$KITTI and MambaPlace$\times$nuScenesFull), columns 19 and 20 are constant 3.0 during feature construction, become identically zero after standardization, and do not change the scorer input; this audit does not extend to the remaining configurations, and it does not imply that the two columns are ineffective in general. The 1-based column indices used in the main text are one greater than the indices listed here.
\end{flushleft}
\end{table}

\section*{S2\quad Complete Exploratory TEST Readings}

Tables~\ref{tabS:cell}--\ref{tabS:counts} contain the complete TEST readings underlying Tables~II and~III of the main paper. Every configuration uses its own frozen Top-25 candidate pool and the Fine predictions already produced for those candidates. Original, Semantic Matching, Rank Fusion, FSR, and SCAR therefore rank the same hypothesis set, and differences isolate the ranking rule. Percentages are reported to two decimals; differences are computed from unrounded readings and then rounded. All TEST comparisons are exploratory because these splits participated in development and analysis. No multiple-comparison correction is applied, and the counts in Table~\ref{tabS:counts} are descriptive point-estimate counts without count-level intervals.

Table~\ref{tabS:cell} reports Cell R@1 for Original, FSR, and SCAR. Table~\ref{tabS:controls} reports the two strong training-free controls, Semantic Matching and Rank Fusion, as Cell R@1 and Top-1@5\,m. For Semantic Matching, Rank Fusion, FSR, and SCAR, Table~\ref{tabS:top1depth} reports Top-1 at 1, 3, and 10\,m, and Table~\ref{tabS:celldepth} reports Cell R@3, R@5, and R@10. Table~IV of the main paper gives their 5\,m and first-ranked cell results. Table~\ref{tabS:originaltop1} gives the Original localization rates at all four distances. Table~\ref{tabS:counts} counts, over configurations, how often the SCAR point estimate is higher than, lower than, or equal to each baseline at Top-1@5\,m.

The Des4Pos rows are the corrected reimplementation described in Section~S4; they are not the historical implementation. The MambaPlace$\times$PandaSet row (marked with a dagger) uses a new ep2 trajectory whose protocol alignment with the other configurations is unresolved; it is a reference reading and is excluded from the aggregate count. After this exclusion, the remaining 24 configurations have SCAR Top-1@5\,m above Original in 23 and Cell R@1 above Original in all 24. The single Top-1 exception is MNCL$\times$PandaSet, where SCAR is $0.15$\,pp below Original; its Cell R@1 is nevertheless $1.80$\,pp above Original. This is the empirical basis of the corresponding statement in the abstract and conclusion.

\begin{table}[H]
\centering
\caption{Complete TEST Cell R@1 (\%). Each cell is the first-ranked cell hit rate under the same fixed Top-25 candidate pool and the same Fine predictions; the three columns differ only in ranking. All TEST readings are exploratory.}
\label{tabS:cell}
\footnotesize
\setlength{\tabcolsep}{4pt}
\begin{tabular}{@{}llccc@{}}
\toprule
Dataset & Backbone & Original & FSR & SCAR \\
\midrule
\multirow{5}{*}{KITTI} & Text2Loc & 29.84 & 46.51 & 43.74 \\
 & Text2Loc++ & 35.00 & 47.51 & 45.02 \\
 & MNCL & 43.90 & 48.19 & 47.41 \\
 & MambaPlace & 24.55 & 44.18 & 40.83 \\
 & Des4Pos (reimpl.) & 23.67 & 44.37 & 43.11 \\
\addlinespace[2pt]
\multirow{5}{*}{PandaSet} & Text2Loc & 45.05 & 43.10 & 52.35 \\
 & Text2Loc++ & 41.15 & 41.35 & 49.10 \\
 & MNCL & 46.05 & 38.95 & 47.85 \\
 & MambaPlace$^{\dagger}$ & 19.80 & 31.75 & 29.50 \\
 & Des4Pos (reimpl.) & 28.10 & 37.30 & 41.55 \\
\addlinespace[2pt]
\multirow{5}{*}{nuScenes} & Text2Loc & 50.61 & 49.01 & 62.19 \\
 & Text2Loc++ & 43.74 & 51.30 & 60.56 \\
 & MNCL & 56.32 & 41.66 & 65.28 \\
 & MambaPlace & 50.29 & 42.46 & 60.11 \\
 & Des4Pos (reimpl.) & 51.63 & 45.40 & 60.11 \\
\addlinespace[2pt]
\multirow{5}{*}{S3DIS} & Text2Loc & 14.39 & 46.49 & 41.70 \\
 & Text2Loc++ & 21.96 & 47.42 & 48.15 \\
 & MNCL & 8.12 & 38.93 & 39.30 \\
 & MambaPlace & 13.47 & 39.30 & 42.25 \\
 & Des4Pos (reimpl.) & 16.05 & 40.41 & 39.11 \\
\addlinespace[2pt]
\multirow{5}{*}{NPM3D} & Text2Loc & 24.14 & 33.62 & 31.03 \\
 & Text2Loc++ & 11.21 & 19.83 & 19.83 \\
 & MNCL & 6.03 & 13.79 & 19.83 \\
 & MambaPlace & 20.69 & 24.14 & 24.14 \\
 & Des4Pos (reimpl.) & 13.79 & 27.59 & 27.59 \\
\bottomrule
\end{tabular}
\end{table}

\begin{table}[H]
\centering
\caption{Complete TEST readings of the two strong training-free controls (\%). Each cell is Cell R@1 / Top-1@5\,m; both controls use the configuration\'s own candidates and Fine predictions.}
\label{tabS:controls}
\footnotesize
\setlength{\tabcolsep}{4pt}
\begin{tabular}{@{}llcccc@{}}
\toprule
Dataset & Backbone & Semantic Matching Cell & Semantic Matching Top-1@5\,m & Rank Fusion Cell & Rank Fusion Top-1@5\,m \\
\midrule
\multirow{5}{*}{KITTI} & Text2Loc & 41.28 & 49.05 & 44.62 & 54.58 \\
 & Text2Loc++ & 44.98 & 56.45 & 46.05 & 59.05 \\
 & MNCL & 46.01 & 54.90 & 46.62 & 57.56 \\
 & MambaPlace & 35.74 & 45.31 & 42.36 & 53.72 \\
 & Des4Pos (reimpl.) & 36.71 & 46.03 & 42.67 & 53.01 \\
\addlinespace[2pt]
\multirow{5}{*}{PandaSet} & Text2Loc & 52.95 & 48.75 & 46.75 & 43.75 \\
 & Text2Loc++ & 47.30 & 42.30 & 44.85 & 41.25 \\
 & MNCL & 34.10 & 30.25 & 38.75 & 35.10 \\
 & MambaPlace$^{\dagger}$ & 28.85 & 19.85 & 30.90 & 22.30 \\
 & Des4Pos (reimpl.) & 40.00 & 29.50 & 41.80 & 32.15 \\
\addlinespace[2pt]
\multirow{5}{*}{nuScenes} & Text2Loc & 58.08 & 50.93 & 54.22 & 47.26 \\
 & Text2Loc++ & 58.72 & 51.20 & 54.60 & 47.21 \\
 & MNCL & 52.74 & 41.36 & 45.90 & 35.54 \\
 & MambaPlace & 53.24 & 41.15 & 47.08 & 36.44 \\
 & Des4Pos (reimpl.) & 55.40 & 37.77 & 50.07 & 34.34 \\
\addlinespace[2pt]
\multirow{5}{*}{S3DIS} & Text2Loc & 29.89 & 28.23 & 38.38 & 36.90 \\
 & Text2Loc++ & 34.32 & 32.10 & 40.41 & 38.56 \\
 & MNCL & 30.26 & 28.97 & 35.79 & 34.69 \\
 & MambaPlace & 31.92 & 29.52 & 37.08 & 35.42 \\
 & Des4Pos (reimpl.) & 33.03 & 30.44 & 36.16 & 34.32 \\
\addlinespace[2pt]
\multirow{5}{*}{NPM3D} & Text2Loc & 33.62 & 27.59 & 33.62 & 27.59 \\
 & Text2Loc++ & 19.83 & 17.24 & 19.83 & 17.24 \\
 & MNCL & 13.79 & 9.48 & 13.79 & 9.48 \\
 & MambaPlace & 24.14 & 19.83 & 24.14 & 19.83 \\
 & Des4Pos (reimpl.) & 27.59 & 23.28 & 27.59 & 23.28 \\
\bottomrule
\end{tabular}
\end{table}

\begin{table}[H]
\centering
\caption{Complete TEST Top-1 localization success (\%) at true-distance thresholds 1, 3, 5, and 10\,m for FSR and SCAR. Both rankings use the same candidates and Fine coordinates; 5\,m is the primary metric.}
\label{tabS:top1fsr}
\footnotesize
\setlength{\tabcolsep}{4pt}
\begin{tabular}{@{}llcccccccc@{}}
\toprule
 & & \multicolumn{4}{c}{FSR} & \multicolumn{4}{c}{SCAR} \\
\cmidrule(lr){3-6}\cmidrule(l){7-10}
Dataset & Backbone & 1\,m & 3\,m & 5\,m & 10\,m & 1\,m & 3\,m & 5\,m & 10\,m \\
\midrule
\multirow{5}{*}{KITTI} & Text2Loc & 6.32 & 36.91 & 58.65 & 74.49 & 6.46 & 37.16 & 59.45 & 75.77 \\
 & Text2Loc++ & 8.13 & 41.01 & 63.36 & 77.59 & 8.29 & 41.23 & 64.15 & 78.86 \\
 & MNCL & 7.34 & 39.26 & 62.00 & 76.54 & 7.20 & 40.40 & 63.82 & 79.69 \\
 & MambaPlace & 7.14 & 36.45 & 57.65 & 73.97 & 7.21 & 36.01 & 57.07 & 72.92 \\
 & Des4Pos (reimpl.) & 6.68 & 36.18 & 57.41 & 72.60 & 6.62 & 36.11 & 57.84 & 73.53 \\
\addlinespace[2pt]
\multirow{5}{*}{PandaSet} & Text2Loc & 6.35 & 29.20 & 40.50 & 49.20 & 7.70 & 30.85 & 44.95 & 55.25 \\
 & Text2Loc++ & 3.10 & 26.10 & 37.90 & 48.10 & 4.55 & 29.45 & 44.30 & 56.30 \\
 & MNCL & 4.50 & 24.10 & 35.40 & 45.10 & 4.65 & 27.20 & 42.05 & 52.50 \\
 & MambaPlace$^{\dagger}$ & 1.60 & 14.90 & 24.15 & 37.95 & 1.80 & 14.35 & 22.15 & 34.65 \\
 & Des4Pos (reimpl.) & 2.85 & 20.70 & 30.45 & 43.35 & 3.25 & 21.60 & 31.55 & 44.65 \\
\addlinespace[2pt]
\multirow{5}{*}{nuScenes} & Text2Loc & 12.47 & 28.07 & 42.58 & 49.13 & 19.10 & 37.67 & 54.68 & 62.43 \\
 & Text2Loc++ & 14.18 & 28.85 & 44.32 & 51.14 & 19.15 & 35.59 & 53.36 & 60.69 \\
 & MNCL & 4.27 & 20.30 & 32.18 & 41.97 & 7.43 & 34.55 & 52.47 & 64.80 \\
 & MambaPlace & 4.04 & 21.06 & 33.32 & 43.14 & 7.24 & 31.68 & 47.71 & 60.21 \\
 & Des4Pos (reimpl.) & 4.24 & 17.14 & 31.03 & 45.39 & 6.27 & 23.79 & 41.44 & 59.76 \\
\addlinespace[2pt]
\multirow{5}{*}{S3DIS} & Text2Loc & 33.95 & 42.25 & 44.83 & 46.31 & 30.26 & 39.30 & 41.33 & 41.70 \\
 & Text2Loc++ & 34.50 & 43.91 & 45.20 & 47.23 & 34.50 & 44.10 & 45.94 & 47.97 \\
 & MNCL & 23.06 & 36.16 & 37.64 & 38.75 & 22.88 & 36.90 & 38.38 & 39.11 \\
 & MambaPlace & 19.93 & 35.06 & 37.45 & 39.11 & 22.14 & 38.19 & 40.77 & 42.07 \\
 & Des4Pos (reimpl.) & 21.77 & 36.90 & 38.38 & 40.22 & 22.32 & 36.35 & 37.82 & 39.11 \\
\addlinespace[2pt]
\multirow{5}{*}{NPM3D} & Text2Loc & 8.62 & 25.00 & 27.59 & 32.76 & 7.76 & 24.14 & 26.72 & 30.17 \\
 & Text2Loc++ & 5.17 & 15.52 & 17.24 & 19.83 & 3.45 & 14.66 & 16.38 & 19.83 \\
 & MNCL & 1.72 & 8.62 & 9.48 & 12.07 & 4.31 & 13.79 & 14.66 & 18.10 \\
 & MambaPlace & 4.31 & 15.52 & 19.83 & 24.14 & 4.31 & 15.52 & 19.83 & 24.14 \\
 & Des4Pos (reimpl.) & 0.00 & 15.52 & 23.28 & 27.59 & 0.00 & 15.52 & 23.28 & 27.59 \\
\bottomrule
\end{tabular}
\end{table}

\begin{sidewaystable}[p]
\centering
\caption{Complete TEST Top-1 localization success (\%) at 1, 3, 5, and 10\,m for Original, Semantic Matching, and Rank Fusion. All three rankings use the same fixed candidate pool and Fine predictions.}
\label{tabS:top1opr}
\small
\setlength{\tabcolsep}{5pt}
\begin{tabular}{@{}llcccccccccccc@{}}
\toprule
 & & \multicolumn{4}{c}{Original} & \multicolumn{4}{c}{Semantic Matching} & \multicolumn{4}{c}{Rank Fusion} \\
\cmidrule(lr){3-6}\cmidrule(lr){7-10}\cmidrule(l){11-14}
Dataset & Backbone & 1\,m & 3\,m & 5\,m & 10\,m & 1\,m & 3\,m & 5\,m & 10\,m & 1\,m & 3\,m & 5\,m & 10\,m \\
\midrule
\multirow{5}{*}{KITTI} & Text2Loc & 3.10 & 20.48 & 35.14 & 48.89 & 4.16 & 28.67 & 49.05 & 66.95 & 5.68 & 33.98 & 54.58 & 70.75 \\
 & Text2Loc++ & 4.51 & 26.14 & 43.72 & 57.77 & 5.84 & 33.66 & 56.45 & 73.16 & 7.48 & 37.54 & 59.05 & 73.68 \\
 & MNCL & 4.36 & 29.14 & 49.59 & 65.70 & 4.89 & 31.85 & 54.90 & 72.96 & 6.40 & 35.90 & 57.56 & 72.86 \\
 & MambaPlace & 3.26 & 18.52 & 31.98 & 44.58 & 4.59 & 26.41 & 45.31 & 62.32 & 6.41 & 33.15 & 53.72 & 69.89 \\
 & Des4Pos (reimpl.) & 2.45 & 17.04 & 29.37 & 41.16 & 4.34 & 27.33 & 46.03 & 62.32 & 5.66 & 32.68 & 53.01 & 68.21 \\
\addlinespace[2pt]
\multirow{5}{*}{PandaSet} & Text2Loc & 7.55 & 28.55 & 40.95 & 50.50 & 7.80 & 35.10 & 48.75 & 59.20 & 7.05 & 31.50 & 43.75 & 52.70 \\
 & Text2Loc++ & 4.20 & 25.45 & 37.35 & 47.45 & 4.35 & 27.80 & 42.30 & 54.90 & 3.55 & 28.60 & 41.25 & 51.30 \\
 & MNCL & 4.95 & 29.30 & 42.20 & 52.60 & 2.50 & 18.95 & 30.25 & 40.70 & 3.90 & 23.50 & 35.10 & 45.30 \\
 & MambaPlace$^{\dagger}$ & 1.35 & 8.70 & 13.90 & 22.80 & 1.55 & 12.10 & 19.85 & 31.20 & 1.70 & 13.20 & 22.30 & 35.40 \\
 & Des4Pos (reimpl.) & 2.65 & 12.65 & 20.35 & 30.00 & 3.05 & 19.35 & 29.50 & 42.50 & 3.20 & 21.80 & 32.15 & 45.55 \\
\addlinespace[2pt]
\multirow{5}{*}{nuScenes} & Text2Loc & 16.56 & 31.31 & 44.63 & 50.80 & 17.29 & 34.55 & 50.93 & 58.19 & 15.12 & 31.74 & 47.26 & 54.38 \\
 & Text2Loc++ & 15.57 & 27.15 & 39.93 & 44.31 & 18.55 & 34.27 & 51.20 & 58.60 & 16.07 & 31.23 & 47.21 & 54.38 \\
 & MNCL & 6.31 & 30.53 & 46.12 & 56.37 & 4.92 & 26.73 & 41.36 & 52.34 & 4.16 & 22.44 & 35.54 & 45.58 \\
 & MambaPlace & 6.75 & 27.30 & 40.67 & 50.55 & 5.89 & 26.88 & 41.15 & 53.22 & 4.47 & 23.22 & 36.44 & 47.39 \\
 & Des4Pos (reimpl.) & 6.01 & 20.55 & 36.03 & 51.48 & 5.30 & 20.98 & 37.77 & 54.91 & 4.78 & 18.63 & 34.34 & 49.57 \\
\addlinespace[2pt]
\multirow{5}{*}{S3DIS} & Text2Loc & 9.41 & 11.99 & 13.84 & 14.39 & 19.37 & 25.83 & 28.23 & 29.70 & 27.86 & 34.69 & 36.90 & 38.19 \\
 & Text2Loc++ & 16.24 & 19.93 & 20.66 & 21.77 & 24.17 & 30.81 & 32.10 & 33.95 & 29.52 & 37.27 & 38.56 & 40.22 \\
 & MNCL & 4.80 & 7.20 & 7.93 & 7.93 & 16.24 & 26.94 & 28.97 & 29.70 & 20.11 & 33.58 & 34.69 & 35.61 \\
 & MambaPlace & 5.90 & 10.52 & 11.99 & 13.47 & 15.31 & 26.94 & 29.52 & 31.55 & 16.05 & 33.21 & 35.42 & 36.90 \\
 & Des4Pos (reimpl.) & 8.12 & 13.84 & 14.58 & 15.87 & 17.34 & 28.60 & 30.44 & 32.66 & 18.63 & 32.66 & 34.32 & 35.98 \\
\addlinespace[2pt]
\multirow{5}{*}{NPM3D} & Text2Loc & 6.03 & 18.10 & 19.83 & 23.28 & 8.62 & 25.00 & 27.59 & 32.76 & 8.62 & 25.00 & 27.59 & 32.76 \\
 & Text2Loc++ & 3.45 & 8.62 & 9.48 & 11.21 & 5.17 & 15.52 & 17.24 & 19.83 & 5.17 & 15.52 & 17.24 & 19.83 \\
 & MNCL & 0.00 & 4.31 & 4.31 & 6.03 & 1.72 & 8.62 & 9.48 & 12.07 & 1.72 & 8.62 & 9.48 & 12.07 \\
 & MambaPlace & 3.45 & 12.93 & 17.24 & 20.69 & 4.31 & 15.52 & 19.83 & 24.14 & 4.31 & 15.52 & 19.83 & 24.14 \\
 & Des4Pos (reimpl.) & 0.00 & 8.62 & 12.93 & 13.79 & 0.00 & 15.52 & 23.28 & 27.59 & 0.00 & 15.52 & 23.28 & 27.59 \\
\bottomrule
\end{tabular}
\end{sidewaystable}

\begin{table}[H]
\centering
\caption{Descriptive counts of SCAR TEST Top-1@5\,m point estimates (higher / lower / tie). Counts use the point estimates in Table~II of the main paper; they carry no count-level confidence interval and are not corrected for multiple comparisons.}
\label{tabS:counts}
\footnotesize
\begin{tabular}{@{}lcc@{}}
\toprule
Comparison & Public backbones (20) & Des4Pos reimpl. (5) \\
\midrule
SCAR vs.\ Original (Top-1@5\,m) & 19 / 1 / 0 & 5 / 0 / 0 \\
SCAR vs.\ FSR & 14 / 5 / 1 & 3 / 1 / 1 \\
SCAR vs.\ strongest rule control & 13 / 6 / 1 & 2 / 2 / 1 \\
\bottomrule
\end{tabular}
\end{table}

\section*{S3\quad Provenance and Reading Rules}

\subsection*{S3.1\quad Checkpoints, trajectories, and completed same-pool results}

The formal SCAR readings used in Tables~II and~III follow localization-metric-first epoch selection. Some configurations did not retain the weights of the selected epoch. For those configurations, the reported result uses a reproduced training trajectory on the frozen TRAIN/VAL features, or weights recovered from the saved per-epoch checkpoints. Weight reuse and reproduced trajectories are labeled separately in the internal record. Backbone, candidate, and Fine caches remain frozen throughout: when the epoch-selection state was completed, only the scorer was trained, and the frozen backbone predictions were reused.

The per-configuration provenance is as follows. Text2Loc++$\times$nuScenes reuses its existing frozen result. Des4Pos$\times$KITTI uses the same-pool result of the corrected test13 run; Des4Pos$\times$S3DIS and Des4Pos$\times$NPM3D use the corrected delivered same-pool results; Des4Pos$\times$PandaSet (seed 0, epoch 4) and Des4Pos$\times$nuScenesFull (seed 0, epoch 16) were completed and accepted in the present round. The latter two cover 2,000 queries over 25 query scene clusters and 7,403 queries over 210 query scene clusters, respectively. Des4Pos$\times$KITTI TEST covers 11,505 queries, 3,932 pose clusters, 4,308 cells, and three scenes; its independent verification passed, and the verification record and file hashes are retained with the delivered assets. MambaPlace$\times$KITTI uses the appended exploratory TEST of the new trajectory at seed 0, epoch 5. MambaPlace$\times$PandaSet uses the new trajectory at epoch 2; its protocol interpretation is unresolved, so it is reported only as a reference.

\subsection*{S3.2\quad Binding of results to caches and code}

Each result is bound to the backbone and scorer checkpoints, the 23-D column order, the TRAIN normalization statistics, candidate identity and order, the Fine cache, the evaluation code, and the cluster definition. Backbone training seeds and scorer initialization seeds are registered separately. The canonical P2 order stored in the cache is the authoritative record. Converting rounded float32 scores to float64 cannot restore the original float64 ordering. One exact-tie boundary sample in Des4Pos$\times$nuScenes once triggered a reconstruction assertion; an order difference is accepted only when all relevant stored ordering columns are exactly tied.

Raw data exports, failed evaluation attempts, and successful evaluations that reuse the same cache are recorded separately. Confidence intervals are bound to the effective computation version and to the resampling unit. Per-query, pose-cluster, and scene-cluster analyses are retained side by side and are not pooled. Scorer-seed repeats estimate initialization variation; backbone training repeats are an independent source of variance and are not interchangeable with scorer-seed repeats.

\subsection*{S3.3\quad Interval construction and reporting}

The five corrected Des4Pos domains each have separate freeze and verification records, and TEST was read once after freezing. KITTI, S3DIS, and NPM3D readings enter the Des4Pos rows of Tables~II and~III; the PandaSet and nuScenesFull readings completed in this round enter the same rows. Paired comparisons use 2,000 bootstrap resamples with seed 123. The statistical unit is the pose, room, or scene cluster registered by each frozen protocol. Intervals are not comparable across different cluster units. Some interval-construction details were not fully preserved in the original computation records. The main text therefore labels an interval as 95\% only when the source states it explicitly; otherwise the stored interval is reported without re-labeling. No multiple-comparison correction is applied.

\section*{S4\quad Reproducibility and Model-Selection Disclosures}

\subsection*{S4.1\quad Model selection}

For every configuration, each epoch is evaluated on the complete validation query set. The formal result uses the localization-metric-first rule: take the state with the highest validation Top-1@5\,m; break exact ties by Cell R@1; break any remaining tie by the earlier epoch. The rule and the seed are fixed for all configurations and are never switched according to TEST readings. The TEST split is not used to select a checkpoint, a threshold, or a configuration for the main tables. Main tables fix backbone training seed 0 and scorer initialization seed 0; other seeds appear only in the controlled VAL studies.

\subsection*{S4.2\quad Des4Pos reimplementation}

The Des4Pos rows are a corrected reimplementation, not the authors' released implementation and not an earlier hybrid implementation used in this project. The historical implementation differs in modules and weight identity, and none of its results enter the main tables. Five early Des4Pos training cells did not enable shuffle in the public commands for either the Coarse or the Fine stage; when the Coarse geometry was flipped, the actual hint direction was not flipped in sync. The corrected version enables TRAIN shuffle and flips the direction words together, and its configuration verification uses TRAIN/VAL only.

The correction was evaluated in the same random seed, the first ten epochs, and the complete KITTI VAL split. Under the Coarse-only diagnostic, Cell R@1 / Cell R@5 is 7.28 / 24.13 for the unaligned setting without shuffle, 13.12 / 36.99 for the aligned setting without shuffle, and 26.61 / 60.40 for the aligned setting with shuffle. These diagnostic readings produce no new Fine or SCAR TEST result and are not part of Tables~II and~III. The improvement is obtained at a single seed and ten epochs, and enabling shuffle changes both batch composition and the random-sampling correspondence. The straight-through estimator, prior injection, and the differences between the paper's L2 objective and the public MSE objective are reimplementation disclosures. The results therefore do not constitute a complete reproduction of the Des4Pos authors' implementation.

\subsection*{S4.3\quad Figure 4 case index}

The three cases in Figure~4 of the main paper come from corrected Des4Pos$\times$KITTI VAL with the frozen formal SCAR head (seed 0, epoch 26). Queries 0002, 0028, and 0071 instantiate the correction, preservation, and harm strata, respectively. Each is the first eligible item in its predefined stratum under the frozen case index; the cases are not a random sample. For query 0002, the first-ranked distance is 5.66\,m under Original and 1.48\,m under SCAR; for query 0028, both are 3.94\,m and the selected cell and point are identical; for query 0071, the distance changes from 3.12\,m to 11.20\,m. Query text is the dataset's English description. GT is used only to annotate the figure and to compute errors; it is never an inference input. The index records query identifiers, scene identifiers, candidate identifiers, and the eligibility rule.

\noindent\textbf{Display convention and query descriptions.}\par
The main-paper panels use east and north coordinates relative to GT. Both axes span $-30$ to $30$\,m in every panel. Orange triangles and purple squares mark the predictions chosen by Original and SCAR, respectively; the markers overlap in the preservation case. Dashed circles have a 5\,m radius. Full query descriptions are reproduced below in their stored order.

\begin{description}
\item[Query 0002.] The pose is east of a gray road. The pose is west of a black wall. The pose is south of a dark-green fence. The pose is on-top of a black fence. The pose is east of a gray sidewalk. The pose is south of a black fence.

\item[Query 0028.] The pose is east of a gray road. The pose is east of a gray sidewalk. The pose is south of a dark-green wall. The pose is east of a black fence. The pose is on-top of a black vegetation. The pose is south of a black terrain.

\item[Query 0071.] The pose is east of a gray road. The pose is north of a black fence. The pose is south of a black vegetation. The pose is west of a black vegetation. The pose is on-top of a black terrain. The pose is east of a gray-green sidewalk.

\end{description}

\subsection*{S4.4\quad Identity of the efficiency row}

In Table~V of the main paper, the row labeled ``Des4Pos (historical impl.)'' reports the earlier hybrid implementation. It is not the corrected reimplementation evaluated in the accuracy tables, and its stage-wise latency must not be read as the cost of the corrected pipeline. The corrected reimplementation was not measured for Table~V. For the other four backbones, the table reports the measured target-domain pipelines; MambaPlace is marked as not measured rather than estimated.

\section*{S5\quad VAL Diagnostics and Controlled Ablations}

All results in this section are development-set (VAL) analyses. They are used to interpret the mechanism and to choose the method; they are not TEST readings and do not enter the main TEST tables. Each study fixes the candidates and Fine predictions and changes only the factor named in its subsection.

\subsection*{S5.1\quad Fixed-pool recoverability and selection risk}

On MNCL$\times$KITTI VAL (3,187 queries, one scene), Original-order Top-1@5\,m is 54.47\%, while 93.22\% of queries have at least one successful Fine prediction in the same Top-25 pool. Of the 1,451 queries that fail under Original order, 1,235 already contain a successful candidate, and 216 have no successful candidate at all. The oracle uses the true position only to identify successful predictions and is an evaluation upper bound, not a deployable rule. Table~\ref{tabS:pool} reports the corrected and broken query counts for two training-free rerankings. FSR corrects 601 and breaks 260 queries, a net gain of 10.70\,pp; a fixed 5\,m consensus rule corrects 219 and breaks 679. Reranking therefore has real recoverable headroom, but an indiscriminate rule can lose more than it gains. These counts motivate a learned scorer that can weigh evidence per candidate rather than apply one fixed radius.

\begin{table}[H]
\centering
\caption{Fixed-pool VAL diagnostics on MNCL$\times$KITTI (3,187 queries, one scene, Top-25). Corrected and broken counts are relative to Original order and are query counts. Intervals elsewhere use 2,000 paired pose-cluster resamples over 1,116 clusters.}
\label{tabS:pool}
\footnotesize
\begin{tabular}{@{}lcccc@{}}
\toprule
Method & Top-1@5\,m (\%) & Corrected & Broken & Note \\
\midrule
Original MNCL & 54.47 & — & — & Original order; 1,451 failures \\
FSR & 65.17 & 601 & 260 & Training-free \\
Fixed 5\,m consensus & 40.04 & 219 & 679 & Naive fixed-radius rule \\
Oracle@25 & 93.22 & — & — & GT upper bound; not deployable \\
\bottomrule
\end{tabular}
\end{table}

\subsection*{S5.2\quad Fine-position readout (F/C/N)}

Table~\ref{tabS:fcnk} reports the F/C/N controls on corrected Des4Pos$\times$KITTI VAL, and Table~\ref{tabS:fcnn} reports the same protocol on MambaPlace$\times$nuScenesFull VAL. F uses the complete 23-D representation with support read at the Fine prediction point. C replaces the support readings in zero-based columns 12--16 with center readings. N masks those columns after normalization, removing Fine support and its derived ranking pathway while retaining center support, center rank, the candidate pool, and all other shared evidence. The three arms share candidates, Fine geometry, labels, normalization, architecture, and epoch selection, so F$-$C and F$-$N isolate the support pathway.

On corrected Des4Pos$\times$KITTI, F reaches 61.09\% Top-1@5\,m and 45.75\% Cell R@1. C reaches 55.32\% and 47.35\%, and N reaches 55.10\% and 46.12\%. The localization gain of F over C is $+5.77$\,pp with interval $[4.42,7.09]$, and over N it is $+5.99$\,pp with interval $[4.68,7.32]$. The Cell R@1 differences point the other way: F$-$C is $-1.60$\,pp with interval $[-3.30,0.19]$, and F$-$N is $-0.38$\,pp with interval $[-2.10,1.32]$. The Fine readout thus improves coordinate precision while not improving, and possibly slightly reducing, first-ranked cell identity. On MambaPlace$\times$nuScenesFull, F reaches 53.32\% Top-1@5\,m and 66.14\% Cell R@1. C is 53.07\% and 66.05\% (F$-$C: $+0.25$\,pp, interval $[-0.10,0.60]$; $+0.09$\,pp, interval $[-0.35,0.55]$), and N is 53.90\% and 66.85\% (F$-$N: $-0.58$\,pp, interval $[-1.48,0.12]$; $-0.71$\,pp, interval $[-1.70,0.08]$). In this configuration the support-pathway contrasts are small and their intervals cross zero. The F$-$FSR rows compare the complete scoring scheme with a training-free rule; they change the ranking rule and do not isolate the support pathway, so they are not support-mechanism estimates.

\begin{table}[H]
\centering
\caption{VAL F/C/N controls on corrected Des4Pos$\times$KITTI (3,187 queries, 1,116 pose clusters, scorer seed 0). F is the complete 23-D feature with support read at the Fine prediction point; C substitutes center readings for the zero-based columns 12--16; N masks those columns after normalization, removing Fine support and the derived ranking pathway while retaining center support (columns 7/8) and center rank. The three arms share candidates, Fine geometry, labels, the normalization pipeline, architecture, and epoch-selection rule. Differences are F minus setting, in percentage points. VAL only; no TEST reading.}
\label{tabS:fcnk}
\footnotesize
\setlength{\tabcolsep}{3pt}
\begin{tabular}{@{}lcccccc@{}}
\toprule
Setting & Center & Fine & Top-1@5\,m & Cell R@1 & Loc.\ diff (95\% CI) & Cell diff (95\% CI) \\
\midrule
F: Fine-position readout & $\times$ & \checkmark & 61.09 & 45.75 & — & — \\
C: Center surrogate & \checkmark & $\times$ & 55.32 & 47.35 & +5.77 [4.42, 7.09] & $-$1.60 [$-$3.30, 0.19] \\
N: Remove cols.\ 12--16 & $\times$ & $\times$ & 55.10 & 46.12 & +5.99 [4.68, 7.32] & $-$0.38 [$-$2.10, 1.32] \\
FSR: Training-free control & --- & --- & 60.59 & 49.58 & +0.50 [$-$0.53, 1.57] & $-$3.83 [$-$5.31, $-$2.38] \\
\bottomrule
\end{tabular}
\end{table}

\begin{table}[H]
\centering
\caption{VAL F/C/N controls on MambaPlace$\times$nuScenesFull (7,635 queries, 209 scene clusters, scorer seed 0). Definitions and shared components follow Table~\ref{tabS:fcnk}. Differences are F minus setting, in percentage points; intervals are paired scene-cluster bootstraps. The FSR row compares the complete scoring scheme with the training-free rule and is not a support-pathway-only contrast. VAL only; no TEST reading.}
\label{tabS:fcnn}
\footnotesize
\setlength{\tabcolsep}{3pt}
\begin{tabular}{@{}lcccccc@{}}
\toprule
Setting & Center & Fine & Top-1@5\,m & Cell R@1 & Loc.\ diff (95\% CI) & Cell diff (95\% CI) \\
\midrule
F: Fine-position readout & $\times$ & \checkmark & 53.32 & 66.14 & — & — \\
C: Center surrogate & \checkmark & $\times$ & 53.07 & 66.05 & +0.25 [$-$0.10, 0.60] & +0.09 [$-$0.35, 0.55] \\
N: Remove cols.\ 12--16 & $\times$ & $\times$ & 53.90 & 66.85 & $-$0.58 [$-$1.48, 0.12] & $-$0.71 [$-$1.70, 0.08] \\
FSR: Training-free control & --- & --- & 39.35 & 48.92 & +13.98 [10.97, 17.01] & +17.22 [13.72, 20.71] \\
\bottomrule
\end{tabular}
\end{table}

\subsection*{S5.3\quad Supervision target: localization success versus GT cell}

On corrected Des4Pos$\times$KITTI VAL, the backbone, Top-25 candidates, Fine coordinates, gate $m\geq4$, and scorer architecture are fixed, and only the definition of a positive candidate changes. Both arms use the same 24,908 TRAIN queries, the same 23-D features, and the same normalization; 8,396 queries have identical positive sets under the two labels. The localization-success label requires the candidate to be in the same scene as the actual Fine position and within 5\,m. The GT-cell label requires the candidate identifier to equal the target cell identifier. Table~\ref{tabS:labels} reports three seeds. The localization label improves Top-1@5\,m by $+2.17$, $+1.51$, and $+1.57$\,pp, with all three intervals above zero. The GT-cell label improves Cell R@1 by $8.60$, $6.46$, and $8.06$\,pp relative to the localization label, again with all intervals excluding zero. The two labels optimize different objects: the localization label directly rewards a first-ranked coordinate within 5\,m, whereas the cell label requires the first rank to fall in the correct map cell. Within this configuration they exhibit a clear trade-off rather than one label improving every metric. The experiment is a VAL development-set analysis and does not change the main protocol's localization-success supervision.

\begin{table}[H]
\centering
\caption{Label-supervision ablation on corrected Des4Pos$\times$KITTI VAL (3,187 queries, 1,116 pose clusters). Backbone, Top-25 candidates, Fine coordinates, gate $m\geq4$, and scorer architecture are fixed; only the positive-label definition changes. Both arms use the same 24,908 TRAIN queries, 23-D features, and normalization pipeline; 8,396 queries have identical positive sets under the two labels. Differences are localization-success label minus GT-cell label, in percentage points; intervals are paired pose-cluster bootstrap 95\% intervals (2,000 resamples, seed 123).}
\label{tabS:labels}
\footnotesize
\setlength{\tabcolsep}{4pt}
\textit{(a) Top-1@5\,m (\%)}\\[2pt]
\begin{tabular}{@{}lcccc@{}}
\toprule
Seed & Localization label & GT-cell label & Difference & 95\% CI \\
\midrule
0 & 60.81 & 58.64 & $+2.17$ & $[0.85, 3.44]$ \\
1 & 60.21 & 58.71 & $+1.51$ & $[0.28, 2.79]$ \\
2 & 60.24 & 58.68 & $+1.57$ & $[0.28, 2.74]$ \\
\bottomrule
\end{tabular}

\medskip
\textit{(b) Cell R@1 (\%)}\\[2pt]
\begin{tabular}{@{}lcccc@{}}
\toprule
Seed & Localization label & GT-cell label & Difference & 95\% CI \\
\midrule
0 & 44.74 & 53.34 & $-8.60$ & $[-10.34, -6.90]$ \\
1 & 46.22 & 52.68 & $-6.46$ & $[-8.14, -4.82]$ \\
2 & 45.06 & 53.12 & $-8.06$ & $[-9.71, -6.46]$ \\
\bottomrule
\end{tabular}
\end{table}

\subsection*{S5.4\quad Directional-gate sensitivity}

On corrected Des4Pos$\times$KITTI VAL, the localization label is fixed and the gate is relaxed from $m\geq4$ to $m\geq1$, $m\geq2$, and $m\geq3$. The $m\geq4$ reconstruction matches the frozen features and the canonical P2 order exactly. Each arm fits its normalization on its own training data and masks columns 19/20 after normalization. Table~\ref{tabS:gate} reports three seeds per gate setting. Top-1@5\,m point differences range from $-0.16$ to $+0.53$\,pp, and every paired interval crosses zero. No stable 5\,m localization gain is observed from relaxing the gate. Cell R@1 shows a small positive direction for $m\geq2$: the interval is above zero for seeds 0 and 1 ($+1.32$ and $+1.29$\,pp) and crosses zero for seed 2 ($+0.09$\,pp). In this candidate set, $m<4$ accounts for 44,537 of 79,675 pairs (55.90\%), so the relaxed gate changes a majority of candidate pairs while producing no consistent localization gain. The main method therefore keeps $m\geq4$.

\begin{table}[H]
\centering
\caption{Directional-gate ablation on corrected Des4Pos$\times$KITTI VAL (3,187 queries, 1,116 pose clusters). The localization label is fixed and the gate is relaxed from $m\geq4$ to $m\geq1,2,3$; 25,961 valid TRAIN queries are used. The $m\geq4$ reconstruction matches the frozen features and P2 exactly; each arm fits normalization on its own training data and masks columns 19/20 after normalization. Differences are the relaxed gate minus $m\geq4$, in percentage points; intervals are paired pose-cluster bootstrap 95\% intervals (2,000 resamples, seed 123).}
\label{tabS:gate}
\footnotesize
\setlength{\tabcolsep}{3pt}
\begin{tabular}{@{}lcccccc@{}}
\toprule
Gate, seed & Gate Top-1@5\,m & $m\geq4$ Top-1@5\,m & Diff (95\% CI) & Gate Cell R@1 & $m\geq4$ Cell R@1 & Diff (95\% CI) \\
\midrule
$m\geq1$, s0 & 60.94 & 61.09 & $-$0.16 [$-$0.96, 0.66] & 45.84 & 45.75 & +0.09 [$-$0.88, 1.00] \\
$m\geq1$, s1 & 60.56 & 60.43 & +0.13 [$-$0.66, 0.97] & 46.22 & 44.56 & +1.66 [0.72, 2.61] \\
$m\geq1$, s2 & 60.87 & 60.34 & +0.53 [$-$0.22, 1.30] & 45.34 & 45.43 & $-$0.09 [$-$1.04, 0.88] \\
\addlinespace[1pt]
$m\geq2$, s0 & 61.28 & 61.09 & +0.19 [$-$0.57, 0.94] & 47.07 & 45.75 & +1.32 [0.44, 2.16] \\
$m\geq2$, s1 & 60.50 & 60.43 & +0.06 [$-$0.66, 0.79] & 45.84 & 44.56 & +1.29 [0.38, 2.22] \\
$m\geq2$, s2 & 60.75 & 60.34 & +0.41 [$-$0.31, 1.17] & 45.53 & 45.43 & +0.09 [$-$0.81, 1.06] \\
\addlinespace[1pt]
$m\geq3$, s0 & 60.97 & 61.09 & $-$0.13 [$-$0.88, 0.66] & 45.09 & 45.75 & $-$0.66 [$-$1.60, 0.19] \\
$m\geq3$, s1 & 60.43 & 60.43 & 0.00 [$-$0.79, 0.75] & 45.40 & 44.56 & +0.85 [$-$0.09, 1.75] \\
$m\geq3$, s2 & 60.65 & 60.34 & +0.31 [$-$0.41, 1.07] & 45.78 & 45.43 & +0.35 [$-$0.51, 1.22] \\
\bottomrule
\end{tabular}
\end{table}

\clearpage
\section*{S6\quad Original Localization Rates}

\begin{table}[!ht]
\centering
\caption{TEST Top-1 localization success (\%) at four distance thresholds for Original.}
\label{tabS:originaltop1}
\small
\begin{tabular*}{\textwidth}{@{\extracolsep{\fill}}llcccc@{}}
\toprule
Dataset & Backbone & 1\,m & 3\,m & 5\,m & 10\,m \\
\midrule
\multirow{5}{*}{KITTI} & Text2Loc & 3.10 & 20.48 & 35.14 & 48.89  \\
 & Text2Loc++ & 4.51 & 26.14 & 43.72 & 57.77  \\
 & MNCL & 4.36 & 29.14 & 49.59 & 65.70  \\
 & MambaPlace & 3.26 & 18.52 & 31.98 & 44.58  \\
 & Des4Pos (reimpl.) & 2.45 & 17.04 & 29.37 & 41.16  \\
\multirow{5}{*}{PandaSet} & Text2Loc & 7.55 & 28.55 & 40.95 & 50.50  \\
 & Text2Loc++ & 4.20 & 25.45 & 37.35 & 47.45  \\
 & MNCL & 4.95 & 29.30 & 42.20 & 52.60  \\
 & MambaPlace$^{\dagger}$ & 1.35 & 8.70 & 13.90 & 22.80  \\
 & Des4Pos (reimpl.) & 2.65 & 12.65 & 20.35 & 30.00  \\
\multirow{5}{*}{nuScenes} & Text2Loc & 16.56 & 31.31 & 44.63 & 50.80  \\
 & Text2Loc++ & 15.57 & 27.15 & 39.93 & 44.31  \\
 & MNCL & 6.31 & 30.53 & 46.12 & 56.37  \\
 & MambaPlace & 6.75 & 27.30 & 40.67 & 50.55  \\
 & Des4Pos (reimpl.) & 6.01 & 20.55 & 36.03 & 51.48  \\
\multirow{5}{*}{S3DIS} & Text2Loc & 9.41 & 11.99 & 13.84 & 14.39  \\
 & Text2Loc++ & 16.24 & 19.93 & 20.66 & 21.77  \\
 & MNCL & 4.80 & 7.20 & 7.93 & 7.93  \\
 & MambaPlace & 5.90 & 10.52 & 11.99 & 13.47  \\
 & Des4Pos (reimpl.) & 8.12 & 13.84 & 14.58 & 15.87  \\
\multirow{5}{*}{NPM3D} & Text2Loc & 6.03 & 18.10 & 19.83 & 23.28  \\
 & Text2Loc++ & 3.45 & 8.62 & 9.48 & 11.21  \\
 & MNCL & 0.00 & 4.31 & 4.31 & 6.03  \\
 & MambaPlace & 3.45 & 12.93 & 17.24 & 20.69  \\
 & Des4Pos (reimpl.) & 0.00 & 8.62 & 12.93 & 13.79  \\
\bottomrule
\end{tabular*}
\par\smallskip
\begin{minipage}{\textwidth}\footnotesize
Des4Pos denotes the corrected reimplementation. All TEST results are exploratory. $^{\dagger}$MambaPlace$\times$PandaSet ep2 is a protocol reference.
\end{minipage}
\end{table}

\clearpage
\newgeometry{left=1.9cm,right=1.9cm,top=1.2cm,bottom=1.2cm,footskip=0.55cm}
\section*{S7\quad Visual Analysis of Localization Hypotheses}
\label{sec:scar-visual-appendix}
\renewcommand{\thefigure}{S\arabic{figure}}
\setcounter{figure}{0}

\subsection*{S7.1\quad Localization Inputs Across Domains}
\begin{figure}[H]
\centering
\makebox[\linewidth][c]{\includegraphics[width=17.2cm]{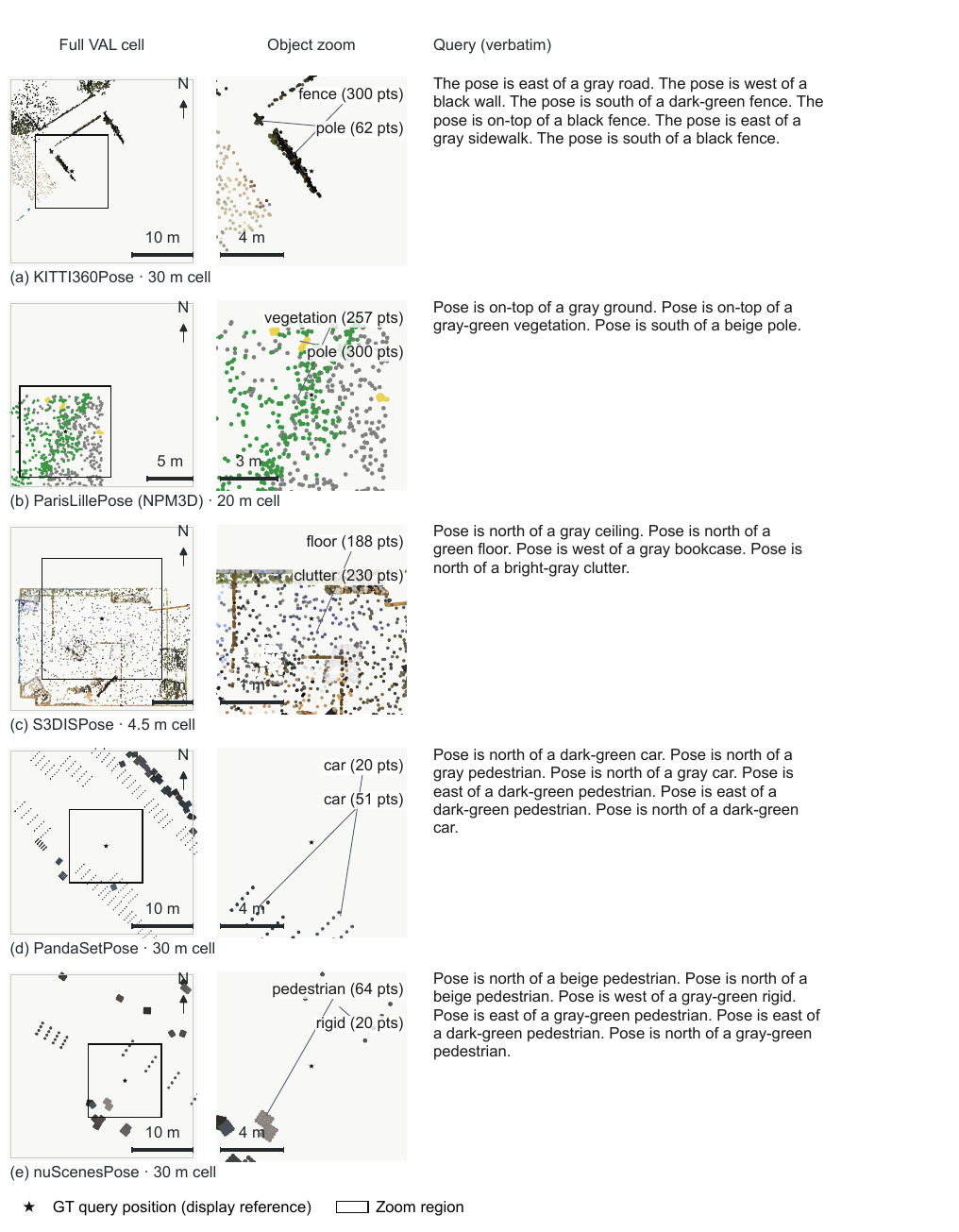}}
\caption{Stored VAL localization cells across five domains. Each row shows a full cell, the boxed region at a larger scale, and its query description. Stars mark GT query positions. Scale bars give the physical extent of each view. The point layers use stored RGB, semantic pseudocolors, or constant object colors as described in the text.}
\label{fig:scar-domain-morphology}
\end{figure}

Figure~\ref{fig:scar-domain-morphology} compares stored localization cells across five domains. PandaSetPose and nuScenesPose use surface samples from annotated boxes with constant object colors. KITTI360Pose and S3DISPose retain RGB; NPM3D uses semantic pseudocolors. Scale bars identify each view's physical extent.

\clearpage
\subsection*{S7.2\quad Correction and Preservation}
The following analysis examines 13 VAL queries using the corrected Des4Pos reimplementation and its frozen seed-0 SCAR heads. Each ranking selects from the same 25 cell--Fine-coordinate hypotheses. A localization hit requires the selected cell to belong to the query scene and its Fine prediction to lie within 5\,m of the ground-truth (GT) query position. Cases are selected by outcome and query identifier to cover correction, preservation, degradation, and joint failure. Their frequencies are not estimates of aggregate performance.

\begin{figure}[H]
\centering
\makebox[\linewidth][c]{\includegraphics[width=17.2cm]{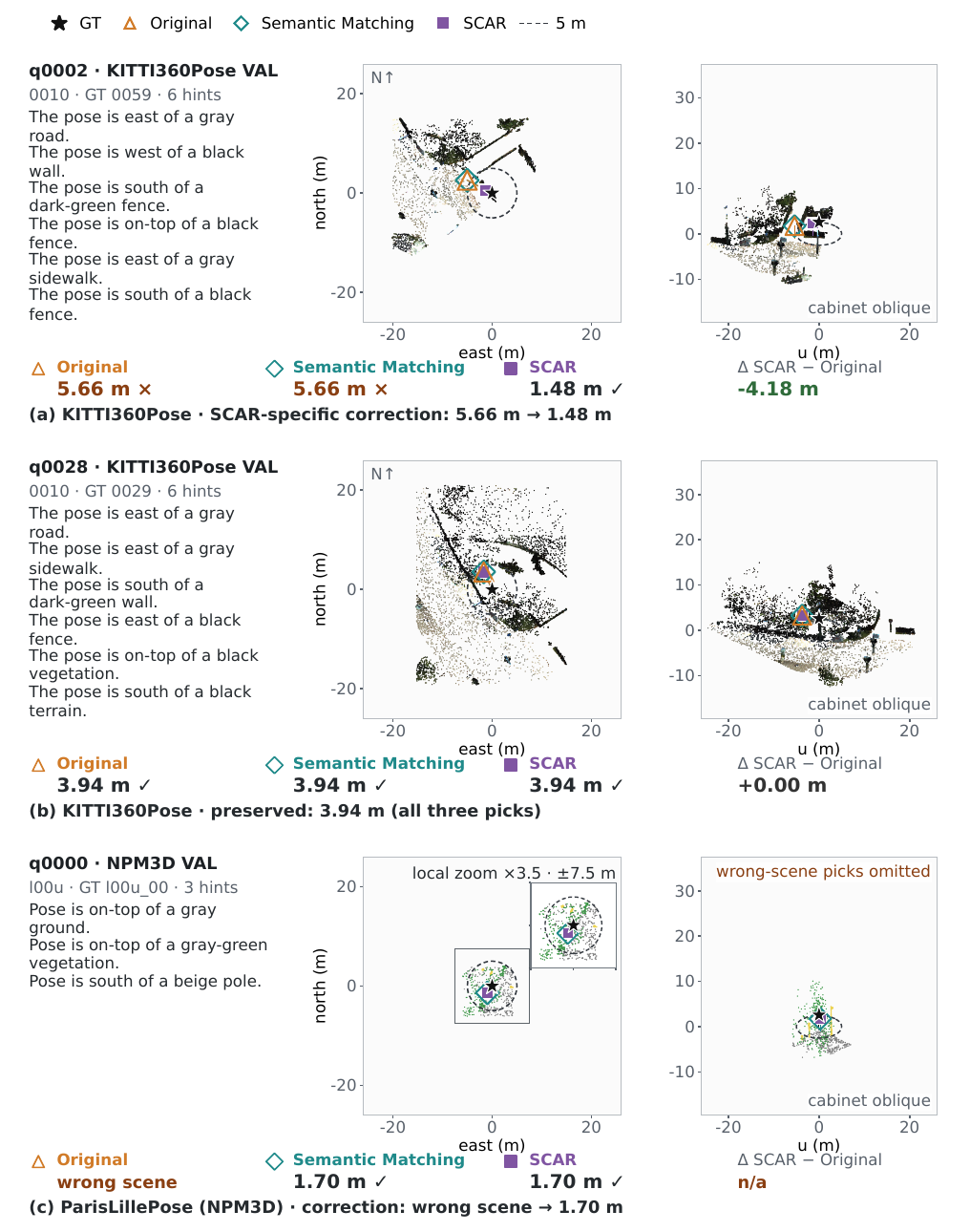}}
\caption{Correction and preservation on KITTI360Pose and NPM3D VAL. (a) SCAR replaces the 5.66\,m prediction selected by both controls with a 1.48\,m prediction. (b) All methods retain the same successful 3.94\,m prediction. (c) Semantic Matching and SCAR replace a wrong-scene selection with the same 1.70\,m prediction. The inset enlarges the boxed region.}
\label{fig:scar-cases-success-a}
\end{figure}
\begin{figure}[H]
\centering
\makebox[\linewidth][c]{\includegraphics[width=17.2cm]{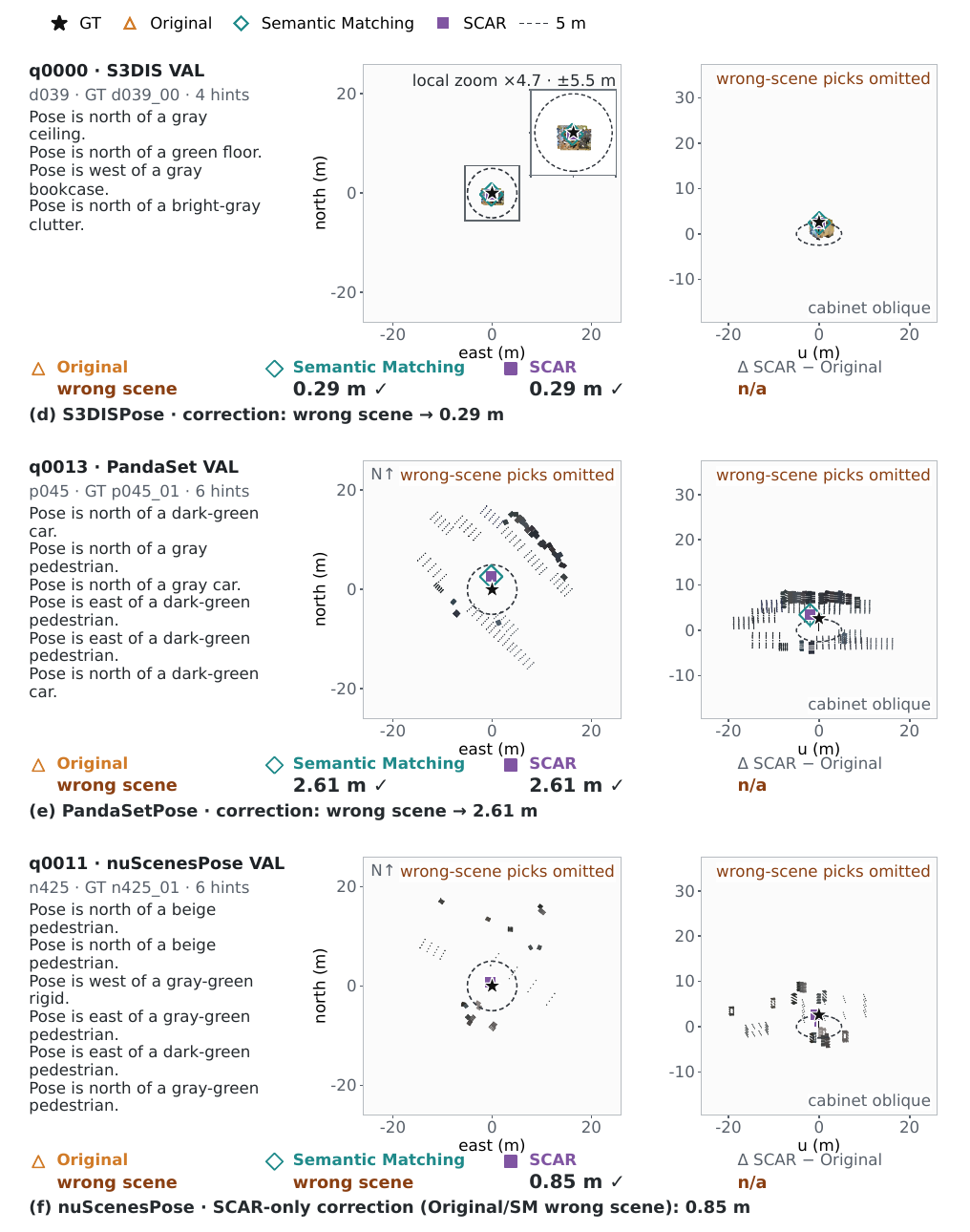}}
\caption{Corrections on S3DISPose, PandaSetPose, and nuScenesPose VAL. (d,e) Semantic Matching and SCAR select the same successful predictions at 0.29 and 2.61\,m. (f) Original and Semantic Matching select wrong-scene cells, while SCAR selects a same-scene prediction at 0.85\,m. The inset enlarges the S3DISPose query region.}
\label{fig:scar-cases-success-b}
\end{figure}

\clearpage
Figures~\ref{fig:scar-cases-success-a} and~\ref{fig:scar-cases-success-b} distinguish corrections shared with semantic matching from changes introduced by SCAR. On KITTI360Pose q0002, SCAR selects the existing 1.48\,m prediction instead of the 5.66\,m prediction chosen by both Original and Semantic Matching. On nuScenesPose q0011, both controls select wrong-scene cells, whereas SCAR selects a same-scene prediction at 0.85\,m. On NPM3D, S3DISPose, and PandaSetPose, Semantic Matching and SCAR select the same successful hypothesis. KITTI360Pose q0028 preserves an unchanged successful prediction at 3.94\,m. The nuScenes examples come from the available 64-query, two-scene VAL subset.

\clearpage
\subsection*{S7.3\quad Same-Scene and Wrong-Scene Failures}
\begin{figure}[H]
\centering
\makebox[\linewidth][c]{\includegraphics[width=17.2cm]{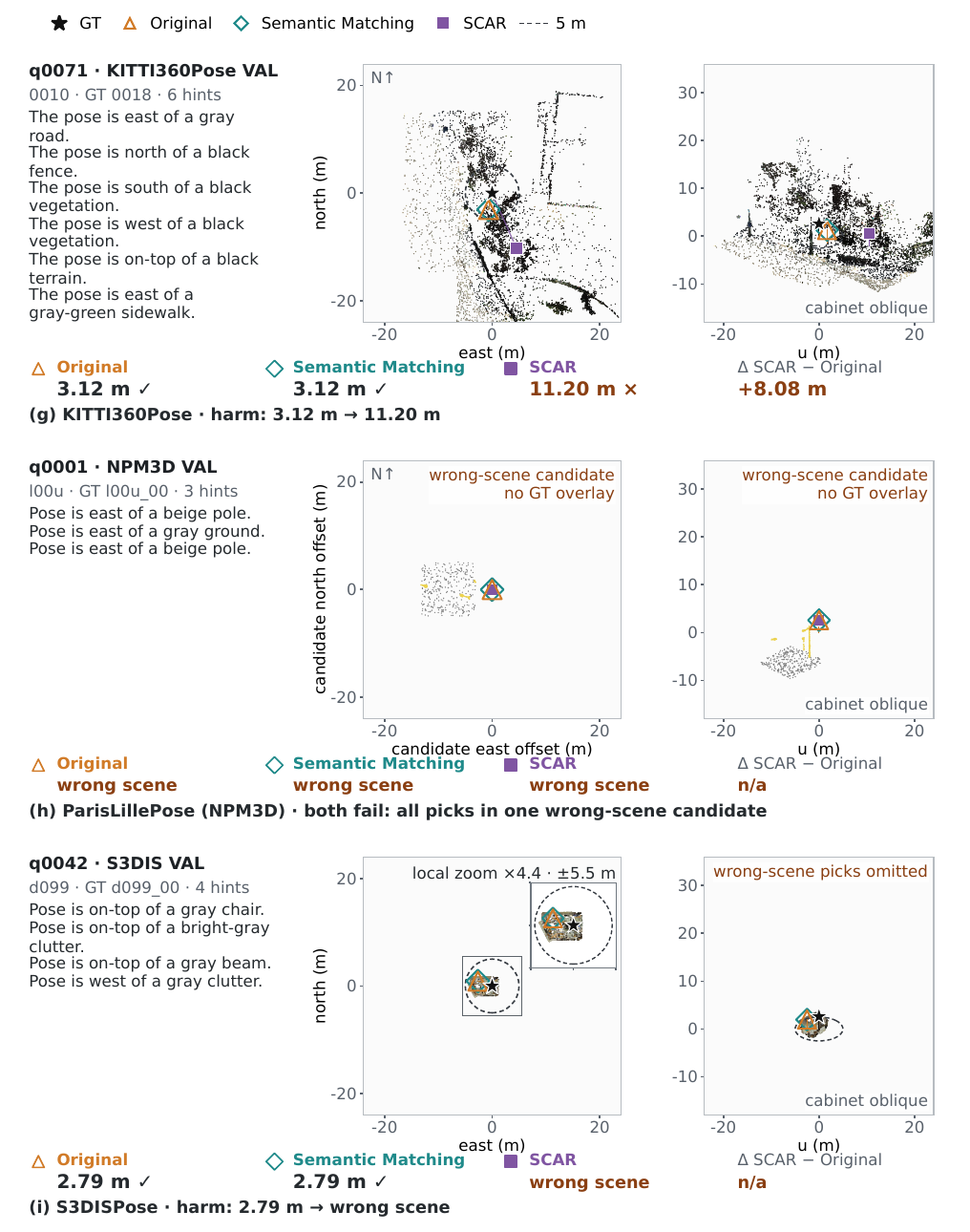}}
\caption{Degradation and joint failure. (g) SCAR replaces a 3.12\,m success with an 11.20\,m prediction. (h) All methods select the same wrong-scene NPM3D cell, shown without a GT overlay. (i) Original and Semantic Matching succeed at 2.79\,m; SCAR selects a wrong-scene cell.}
\label{fig:scar-cases-failure-a}
\end{figure}

\begin{figure}[H]
\centering
\makebox[\linewidth][c]{\includegraphics[width=17.2cm]{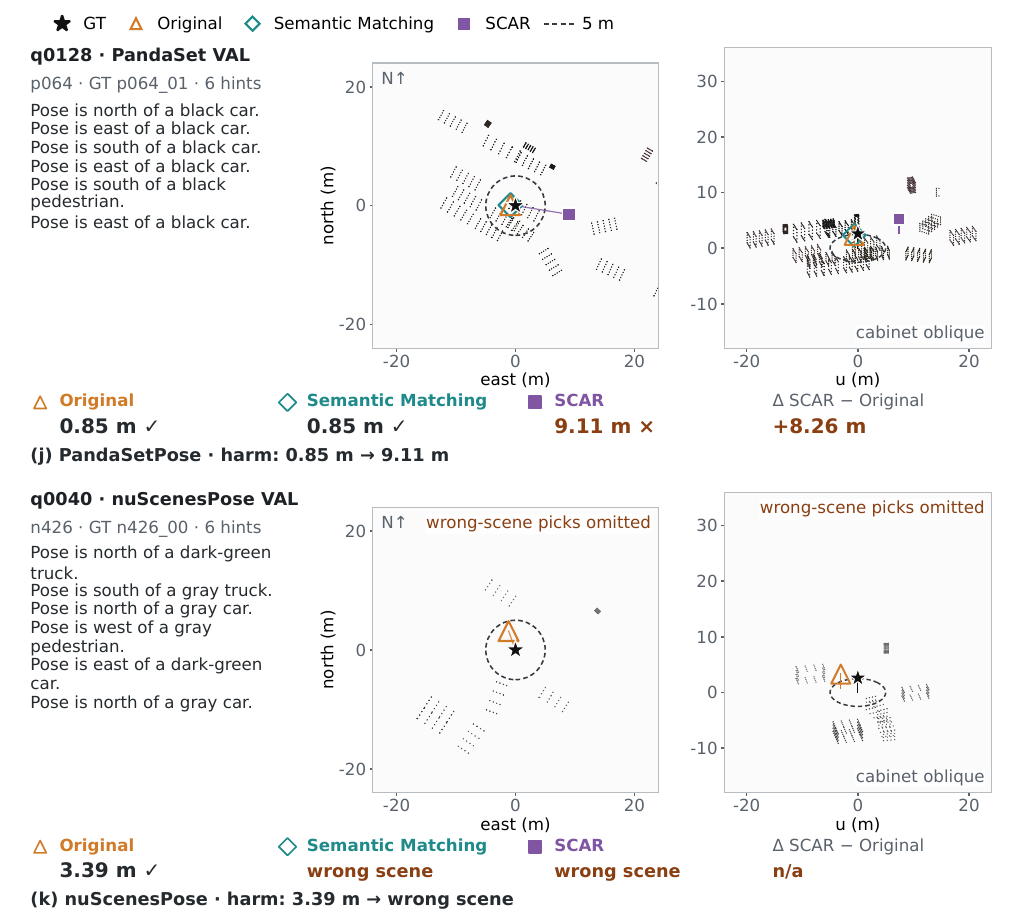}}
\caption{Degradation on PandaSetPose and nuScenesPose VAL. (j) On PandaSetPose q0128, SCAR replaces the 0.85\,m prediction selected by Original and Semantic Matching with a same-scene prediction at 9.11\,m. (k) On nuScenesPose q0040, Original selects a successful prediction at 3.39\,m, while Semantic Matching and SCAR select the same wrong-scene cell. Wrong-scene predictions have no finite reported error and are excluded from the GT-frame overlays.}
\label{fig:scar-cases-failure-b}
\end{figure}

Figures~\ref{fig:scar-cases-failure-a} and~\ref{fig:scar-cases-failure-b} distinguish inaccurate same-scene predictions from wrong-scene selections. The NPM3D example concerns three failed first selections; whether another successful hypothesis exists in its Top-25 pool is a separate question.

\clearpage
\clearpage
\subsection*{S7.4\quad Candidate-Ranking Paths}
\begin{figure}[H]
\centering
\makebox[\linewidth][c]{\includegraphics[width=17.2cm]{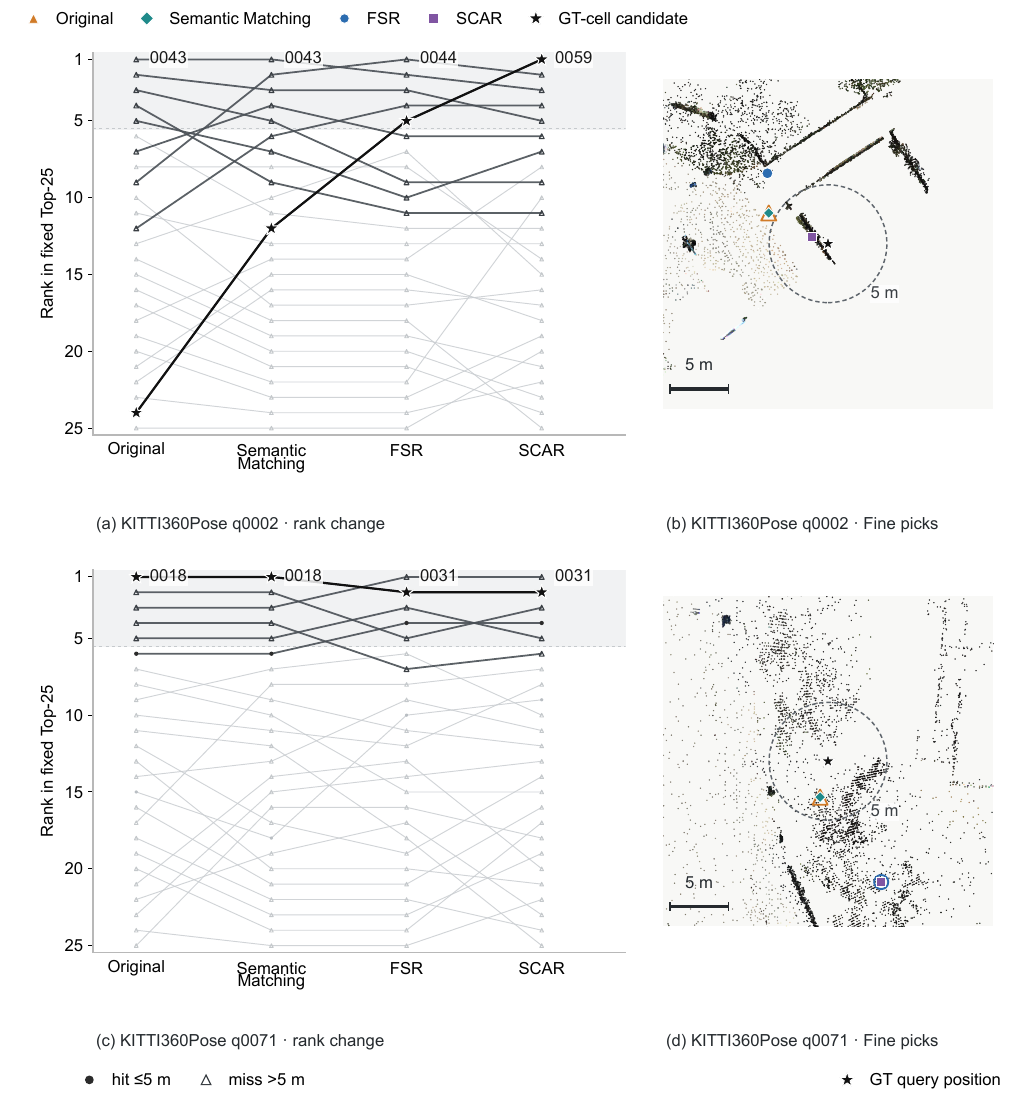}}
\caption{Candidate-ranking paths on corrected Des4Pos$\times$KITTI VAL. (a,c) Each line tracks one fixed candidate across four rankings; candidates appearing in any Top-5 are highlighted. Stars track the GT-cell candidate. (b,d) Existing Fine predictions of the first-ranked candidates; stars mark GT query positions. Candidate identity and coordinate accuracy are separate criteria. Both pools contain only same-scene candidates.}
\label{fig:scar-ranking-paths}
\end{figure}

Figure~\ref{fig:scar-ranking-paths} traces candidate identities through four rankings of the same pool. For KITTI360Pose q0002, the GT-cell candidate moves from rank 24 in Original to ranks 12, 5, and 1 under Semantic Matching, FSR, and SCAR. Its existing Fine prediction is the only one within 5\,m in this pool. For q0071, FSR and SCAR both promote the 11.20\,m prediction above the 3.12\,m GT-cell prediction. The candidate paths explain how a change in ordering alters the output without changing any predicted coordinate.

\clearpage
\subsection*{S7.5\quad Evidence Attached to Selected Hypotheses}
\begin{figure}[H]
\centering
\makebox[\linewidth][c]{\includegraphics[width=17.2cm]{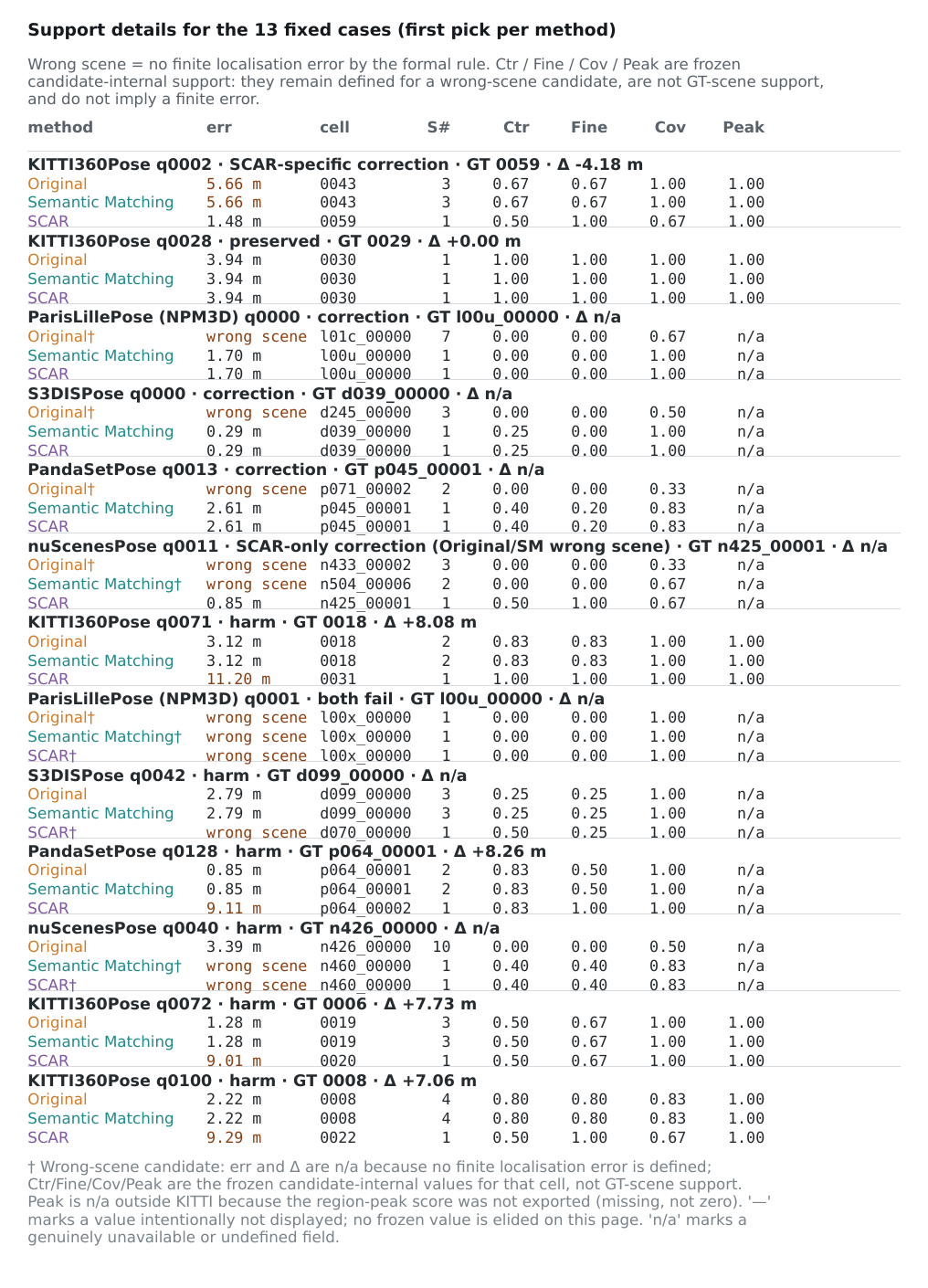}}
\caption{Selected-hypothesis evidence for 13 VAL cases and three rankings. Columns report the cell identity, SCAR rank, localization error, center support (Ctr), Fine-position support (Fine), matching coverage (Cov), and regional peak support (Peak). Wrong-scene selections fail localization but retain their candidate-internal support. Peak is n/a outside KITTI because it was not exported.}
\label{fig:scar-support}
\end{figure}
\clearpage
\subsection*{S7.6\quad Support and Coordinate Accuracy}
\begin{figure}[H]
\centering
\makebox[\linewidth][c]{\includegraphics[width=17.2cm]{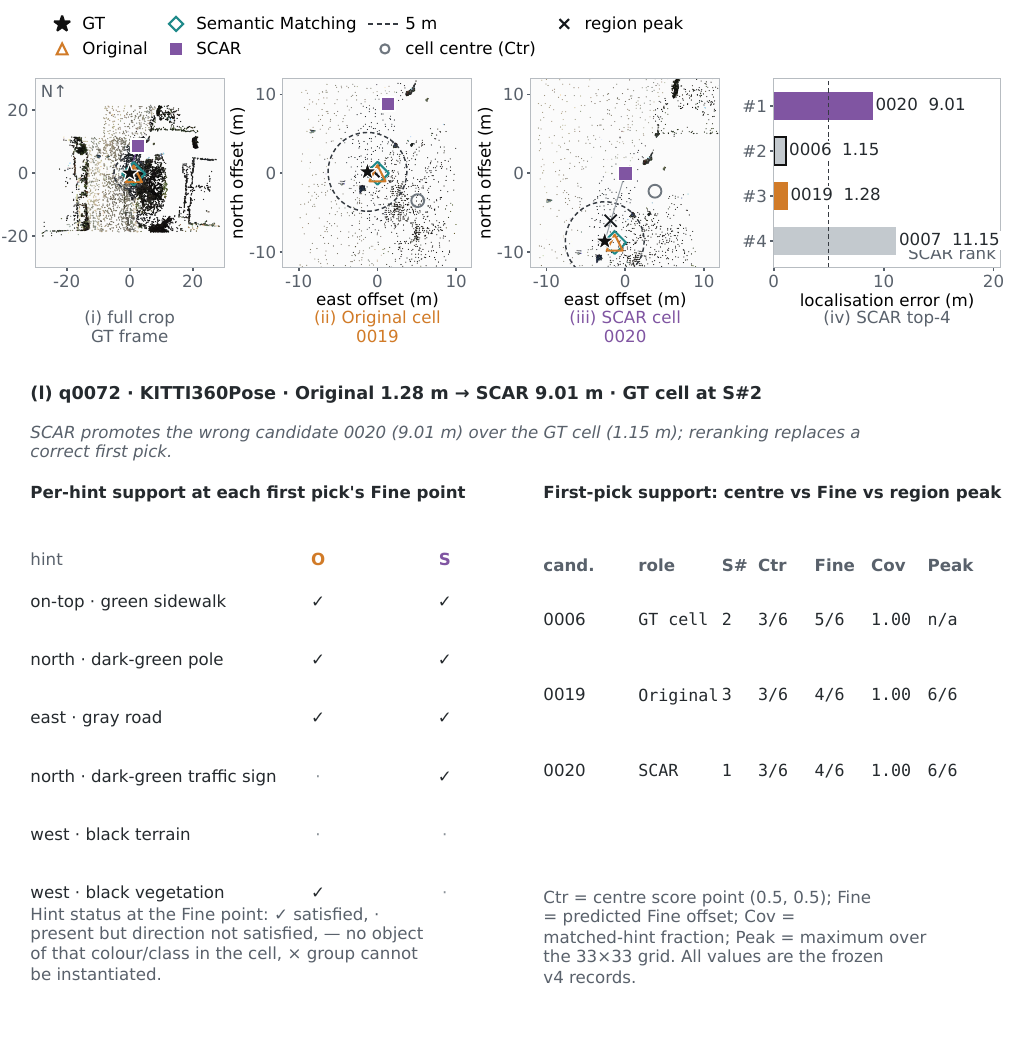}}
\caption{KITTI360Pose q0072: a ranking failure despite matching support summaries. (i) GT-frame overview. (ii,iii) Original and SCAR candidate cells. (iv) Fine errors of the SCAR Top-4 candidates; the dashed line marks 5\,m. SCAR ranks the 9.01\,m prediction first, ahead of the GT-cell candidate at 1.15\,m, while Original selects a prediction at 1.28\,m. The tables report per-hint outcomes and candidate-internal support. Cell identity and Fine localization error are evaluated separately.}
\label{fig:scar-evidence-q72}
\end{figure}

The detailed KITTI cases connect the selected coordinates to their support summaries. On q0072 (Fig.~\ref{fig:scar-evidence-q72}), Original selects a prediction at 1.28\,m. SCAR instead ranks a 9.01\,m prediction first and leaves the GT-cell candidate, whose Fine error is 1.15\,m, second. The Original and SCAR selections have identical stored center support, Fine support, matching coverage, and regional peak. These summaries alone therefore do not distinguish the two selections.

\begin{figure}[H]
\centering
\makebox[\linewidth][c]{\includegraphics[width=17.2cm]{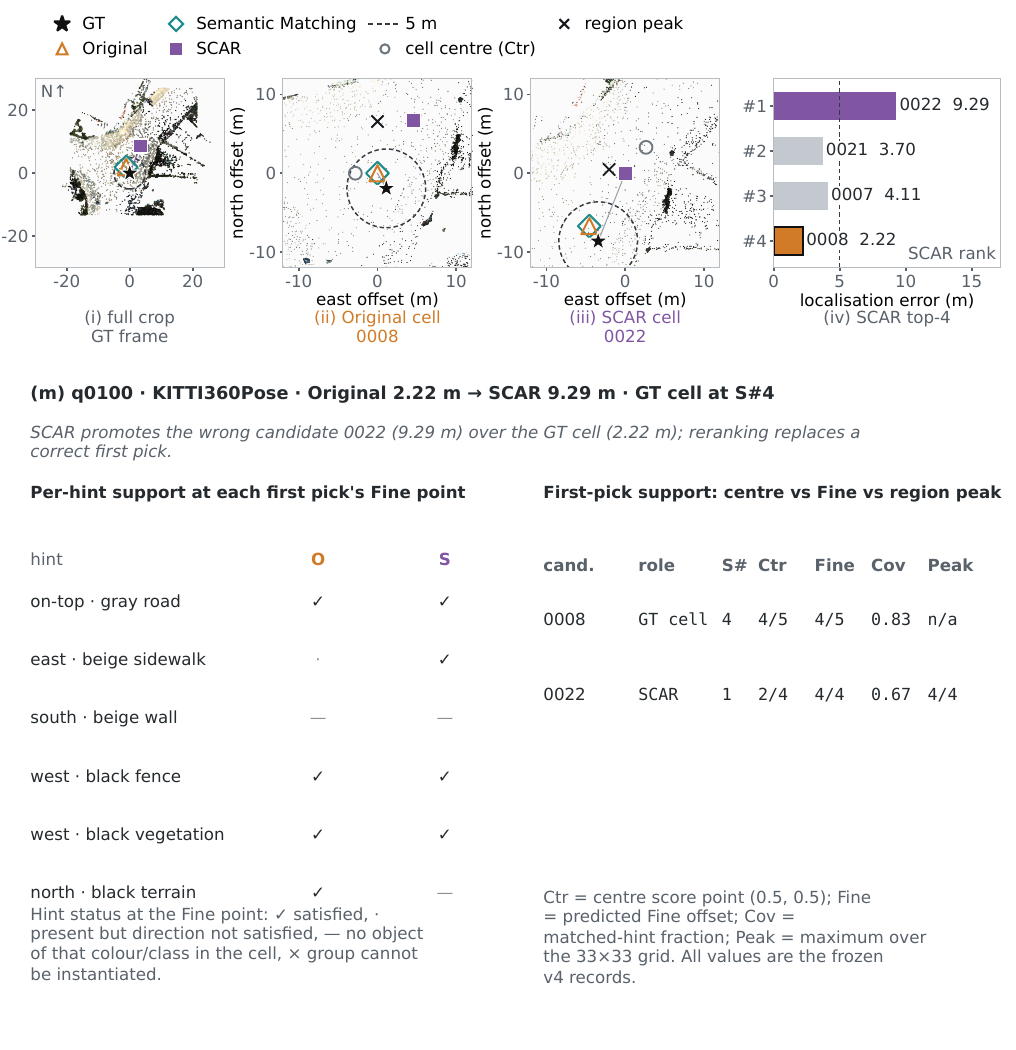}}
\caption{KITTI360Pose q0100: high Fine support with an inaccurate selected coordinate. Panels and tables follow Fig.~\ref{fig:scar-evidence-q72}. SCAR promotes the 9.29\,m prediction, while the Original prediction at 2.22\,m falls to SCAR rank 4. The SCAR selection has Fine support 1.00 and matching coverage 0.67; the Original selection has Fine support 0.80 and coverage 0.83. Support is evaluated among eligible matched relations, whereas localization error is measured against the GT query position.}
\label{fig:scar-evidence-q100}
\end{figure}

\clearpage
On q0100 (Fig.~\ref{fig:scar-evidence-q100}), SCAR selects a 9.29\,m prediction with Fine support 1.00, while Original selects a 2.22\,m prediction with Fine support 0.80. Matching coverage is lower for the SCAR selection, 0.67 versus 0.83. High support among eligible matched relations can thus coexist with a larger localization error. Joint inspection of support and coverage explains this ambiguity; the controlled F/C/N comparisons in Section~S5.2 quantify the effect of changing the support pathway.

\clearpage
\subsection*{S7.7\quad Cross-Configuration Localization and Retrieval Gains}
\begin{figure}[H]
\centering
\makebox[\linewidth][c]{\includegraphics[width=17.2cm]{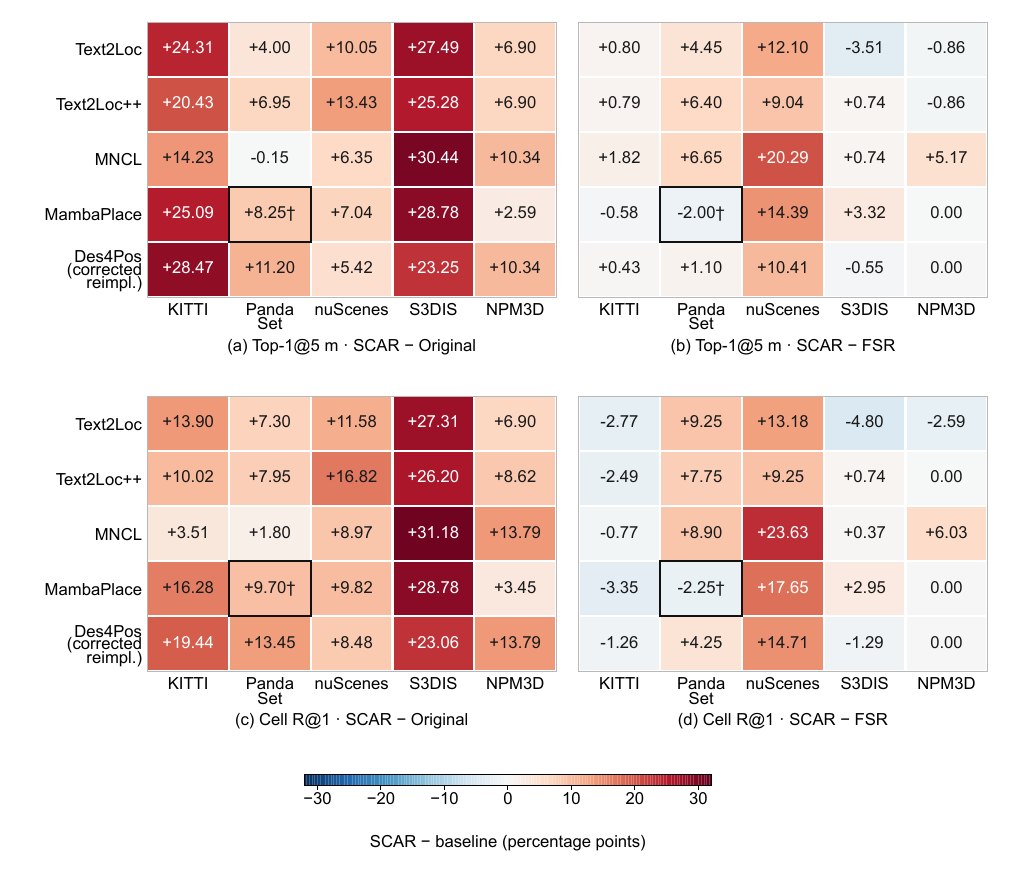}}
\caption{SCAR gains across five backbones and five domains. (a,b) Top-1@5\,m differences from Original and FSR. (c,d) Corresponding Cell R@1 differences. Entries report SCAR minus the indicated baseline in percentage points, using unrounded source values. All panels share a zero-centered color scale spanning $[-32,32]$\,pp. Colors encode point estimates, not statistical significance. The outlined MambaPlace$\times$PandaSet configuration is a protocol reference; Des4Pos denotes the corrected reimplementation. All TEST comparisons are exploratory.}
\label{fig:scar-gain-map}
\end{figure}

Figure~\ref{fig:scar-gain-map} summarizes exploratory TEST point estimates for five backbones across five domains. Relative to Original, SCAR improves Top-1@5\,m in 24 configurations and Cell R@1 in all 25. Relative to FSR, the positive/negative/tied counts are 17/6/2 for Top-1@5\,m and 13/9/3 for Cell R@1. The KITTI rows illustrate that a higher localization success rate can accompany lower first-ranked cell recall. The outlined MambaPlace$\times$PandaSet row is included as a protocol reference; excluding it leaves the 24 configurations used for the abstract's aggregate statement. The heatmap visualizes the existing results in Section~S2; its color scale encodes percentage-point differences.

